\documentclass[sigconf,screen]{acmart}

\usepackage{multirow}
\usepackage{colortab}
\usepackage{colortbl}
\newcommand{\bhline}[1]{\noalign{\hrule height #1}}   
\usepackage{threeparttable}
\usepackage{arydshln}
\usepackage{setspace}
\usepackage{color}
\usepackage{bm}
\usepackage{algorithm}
\usepackage[noend]{algorithmic}
\usepackage[subrefformat=parens]{subcaption}

\AtBeginDocument{%
  \providecommand\BibTeX{{%
    \normalfont B\kern-0.5em{\scshape i\kern-0.25em b}\kern-0.8em\TeX}}}

\copyrightyear{2023} 
\acmYear{2023} 
\setcopyright{acmlicensed}\acmConference[GECCO '23]{Genetic and Evolutionary Computation Conference}{July 15--19, 2023}{Lisbon, Portugal}
\acmBooktitle{Genetic and Evolutionary Computation Conference (GECCO '23), July 15--19, 2023, Lisbon, Portugal}
\acmPrice{15.00}
\acmDOI{10.1145/3583131.3590360}
\acmISBN{979-8-4007-0119-1/23/07}

\acmSubmissionID{pap168}

\begin{document}

\title{Fuzzy-UCS Revisited: Self-Adaptation of Rule Representations in Michigan-Style Learning Fuzzy-Classifier Systems}


\author{Hiroki Shiraishi}
\orcid{0000-0001-8730-1276}
\email{shiraishi-hiroki-yw@ynu.jp}
\affiliation{%
  \institution{Department of Electrical Engineering and Computer Science\\Yokohama National University}
  \city{Yokohama}
  \country{Japan}
  }
\author{Yohei Hayamizu}
\orcid{0000-0003-1642-4919}
\email{yhayami1@binghamton.edu}
\affiliation{%
  \institution{Department of Computer Science\\The State University of New York at Binghamton}
  \city{Binghamton}
  \country{United States}
  }
\author{Tomonori Hashiyama}
\orcid{0000-0001-7218-2999}
\email{hashiyama.tomonori@uec.ac.jp}
\affiliation{%
  \institution{Department of Informatics\\ The University of Electro-Communications}
  \city{Tokyo}
  \country{Japan}
  }


\begin{abstract}

This paper focuses on the impact of rule representation in Michigan-style \textit{Learning Fuzzy-Classifier Systems} (LFCSs) on its classification performance. A well-representation of the rules in an LFCS is crucial for improving its performance. However, conventional rule representations frequently need help addressing problems with unknown data characteristics. To address this issue, this paper proposes a supervised LFCS (i.e., Fuzzy-UCS) with a self-adaptive rule representation mechanism, entitled Adaptive-UCS. Adaptive-UCS incorporates a \textit{fuzzy indicator} as a new rule parameter that sets the membership function of a rule as either rectangular (i.e., crisp) or triangular (i.e., fuzzy) shapes. The fuzzy indicator is optimized with evolutionary operators, allowing the system to search for an optimal rule representation. 
Results from extensive experiments conducted on continuous space problems demonstrate that Adaptive-UCS outperforms other UCSs with conventional crisp-hyperrectangular and fuzzy-hypertrapezoidal rule representations in classification accuracy. Additionally, Adaptive-UCS exhibits robustness in the case of noisy inputs and real-world problems with inherent uncertainty, such as missing values, leading to stable classification performance.

\end{abstract}




\maketitle
\section{Introduction}

Michigan-style \textit{Learning Classifier Systems} (LCSs) \cite{holland1986possibilities} are a paradigm of evolutionary machine learning that utilizes genetic algorithms (GAs) \cite{goldberg1989genetic} to generate a set of accurate and general IF-THEN rules (known as \textit{classifiers}). 
To date, numerous extensions to LCS have been proposed \cite{urbanowicz2009learning}, with the majority of them based on the most prominent LCS - the \textit{XCS Classifier System} (XCS) \cite{wilson1995xcs}.
One of the XCS extensions, the \textit{sUpervised Classifier System} (UCS) \cite{bernado2003accuracy}, learns the rules in a supervised learning fashion.
As a result, 
UCS has been frequently employed as a data analysis technique \cite{urbanowicz2015exstracs,zhang2021lcs} because it is suitable for classification problems such as the analysis of genetic heterogeneity in bladder cancer \cite{urbanowicz2013role} and the movement analysis of rehabilitation exercises for post-operative patients \cite{guevara2021intelligent}.

LCS, including UCS, aims to establish a ruleset that comprehensively covers the entire problem space so that each rule makes sound decisions \cite{shiraishi2022can}.
The goal of LCS is to attain an accurate and concise ruleset for a given problem \cite{shoeleh2015knowledge}.
Thus, a rule representation setup is important to effectively cover the problem space with a minimal number of rules, depending on the nature of the problem. For instance, the crisp-hyperrectangular representation \cite{stone2003real} is the most commonly introduced rule representation for addressing real-valued input problems. The crisp-hyperrectangular representation characterizes the rule condition, which corresponds to the antecedent (IF) part of a rule, as an interval-based hyperrectangle with which UCS determines whether an input matches a rule. 
However, 
this representation is insufficient in generating accurate rules for problem spaces with class boundaries that cannot be partitioned by hyperrectangles, such as diagonals or curves, or for problem spaces with vague class boundaries, such as real-world data sets, leading to subpar classification performance \cite{butz2008function,shiraishi2022can}.

To address this issue, Orriols-Puig et al. \cite{orriols2008fuzzy,orriols2008evolving} proposed the \textit{sUpervised Fuzzy-Classifier System} (Fuzzy-UCS).
Fuzzy-UCS integrates \textit{fuzzy logic} \cite{zadeh1965fuzzy,zadeh1973outline} into UCS. 
Crisp rules use binary logic, either \textit{true} (1) or \textit{false} (0), to express the matching degree between an input and a rule, while fuzzy rules use continuous values ranging from 0 to 1 by fuzzy logic.
Fuzzy-UCS focuses on the robustness of fuzzy logic in handling noisy input and producing output classes with a degree of certainty in classification problems \cite{casillas2007fuzzy,shoeleh2015knowledge}. The authors aimed to solve various classification problems using fuzzy rules with triangular-shaped membership functions serving as the fuzzy set of rule conditions.
However, their experiments revealed that Fuzzy-UCS was superior to UCS in only 8 out of 20 real-world data problems, and inferior in the remaining 12 problems in classification performance \cite{orriols2008fuzzy}.
This highlights the challenge of achieving high classification performance in the current (Fuzzy-)UCS framework, which operates exclusively on either fuzzy or crisp rules.

To take the most advantage of fuzzy logic without declining classification performance,
we propose a\textit{ sUpervised Self-Adaptive-Classifier System} (Adaptive-UCS), a Fuzzy-UCS with a self-adaptive rule representation mechanism.
 Adaptive-UCS incorporates a \textit{fuzzy indicator} parameter, which governs the shape of the membership function of a rule's fuzzy set as either crisp or fuzzy. 
 This study employs simple rectangular- and triangular-shaped membership functions. The fuzzy indicator enables the system to seamlessly operate with both crisp-hyperrectangular and fuzzy hypertriangular rule representations. Optimizing the fuzzy indicators is achieved through evolutionary operators (i.e., crossover and mutation), which allows the system to search for the optimal representation for a given problem. 
 The optimization process eliminates the need for a preliminary setup for rule representation that is typically required in UCS and Fuzzy-UCS when encountering unknown problems.

The organization of the paper is outlined as follows: 
Sect. \ref{sec: related work} provides an overview of relevant literature on rule representation in LCSs. 
Sect. \ref{sec: Fuzzy-UCS} outlines Fuzzy-UCS.
Sect. \ref{sec: proposed} describes the proposed system, Adaptive-UCS.
In Sects. \ref{sec: experiment 1} and \ref{sec: experiment 2}, comparative experiments are conducted using three benchmark problems and 20 real-world dataset classification problems, respectively.
These experiments compare the performance of UCS using crisp-hyperrectangular rule representation \cite{stone2003real}, Fuzzy-UCS using fuzzy-hypertrapezoidal rule representation \cite{shoeleh2011towards} (described later in Sect. \ref{sec: related work}), and Adaptive-UCS.
The results of the experiments are then evaluated and discussed. 
Finally, Sect. \ref{sec: concluding remarks} concludes the paper and highlights its contributions and potential avenues for future work.

\section{Related Work}
\label{sec: related work}

This section provides an overview of the various crisp and fuzzy rule representations that have been proposed in the context of LCSs that handle real-valued input. 
It is worth noting that 
LCSs that utilize fuzzy rules are commonly referred to as \textit{Learning Fuzzy-Classifier Systems} (LFCSs) \cite{bonarini1999introduction,valenzuela1991fuzzy}.

The hyperrectangular representation \cite{wilson1999xcsr,wilson2000mining,stone2003real,dam2005real}, despite being among the first proposed, is considered the most prevalent crisp rule representation in data analysis \cite{bishop2020optimality}.  
Its straightforward and highly legible rule structure makes data easily interpretable and analyzable by humans \cite{heider2022investigating}.
Subsequently, more complex crisp rule representations such as hyperellipsoid \cite{butz2008function}, convex hull \cite{lanzi2006using}, gene expression programming \cite{wilson2008classifier}, code fragment \cite{iqbal2013reusing,arif2017solving}, multi-layer perceptron neural network \cite{bull2002accuracy}, and curved surface hyperpolyhedron \cite{shiraishi2022beta} have been proposed in order to more accurately approximate complex class boundaries.
Recently, Shiraishi et al. \cite{shiraishi2022can} proposed a mechanism for utilizing both hyperrectangular and curved surface hyperpolyhedral rules within a single XCS system. 

The seminal works of the prominent Michigan-style LFCS, namely Fuzzy-XCS \cite{casillas2007fuzzy} and Fuzzy-UCS \cite{orriols2008fuzzy}, employed a fuzzy rule representation based on triangular-shaped membership functions. This representation has since become widely adopted in numerous works (e.g., \cite{orriols2008approximate,orriols2011fuzzy,chen2016accuracy,bishop2021genetic}).
Bull and O'Hara \cite{bull2002accuracy} presented a neuro-fuzzy classifier system utilizing fuzzy radial basis function neural networks and evaluated its performance in function prediction problems. 
Tadokoro et al. \cite{tadokoro2021xcs} introduced fuzzy rules with probability density functions of the multivariate normal distribution (MVN) as the membership function and demonstrated its effectiveness in classification problems in that real-world data conform to an MVN.

Shoeleh et al. \cite{shoeleh2010handle,shoeleh2011towards} 
proposed an XCS-like LFCS system that utilizes rule conditions represented by trapezoidal-shaped membership functions (termed the hypertrapezoidal representation). 
By doubling the number of condition parameters (i.e., changing the membership function from a 2-variable rectangle to a 4-variable trapezoid) of a hyperrectangular rule that expresses only certain regions, they made it possible to express both \textit{certain} regions, characterized by a matching degree of 1, and \textit{vague} regions, characterized by a matching degree in the range of (0,1), simultaneously in a single rule.
However, an increase in the number of parameters in the rule condition may result in an expanded search space, thus potentially hindering system performance, as evidenced by the experimental results of Lanzi and Wilson \cite{lanzi2006using}, which showed a decrease in performance with an increase in the number of vertices in the convex hull representation\footnote{Lanzi and Wilson \cite{lanzi2006using} evaluated the performance of convex hull representations with varying numbers of vertices in the convex hull, namely 3, 5, 10, and 15, on three 2D function prediction problems. The results revealed that the representation with the lowest (highest) expressiveness, i.e., 3 (15) points, exhibited the best (worst) function prediction performance for all problems.}.
Additionally, the hypertrapezoidal representation has been found to perform inferiorly in some real-world data classification problems compared to the hyperrectangular representation (cf. \cite{shoeleh2011towards}).
Therefore, in this paper, we propose a self-adaptive methodology in which a membership function with a reduced number of variables (i.e., two) is adaptively adjusted for each rule, thus allowing a single ruleset in a Fuzzy-UCS system to represent both \textit{certain} and \textit{vague} rule-covering regions.

\section{Fuzzy-UCS in a Nutshell}
\label{sec: Fuzzy-UCS}
The \textit{sUpervised Fuzzy-Classifier System} (Fuzzy-UCS) \cite{orriols2008fuzzy} is an LFCS that combines rule learning and genetic algorithms (GAs) to evolve fuzzy rules online. 
The system operates in two distinct modes: training (\textit{exploration}) and test (\textit{exploitation}). Fuzzy-UCS searches for accurate and maximally general rules in the training mode and utilizes the acquired rules to infer classes from unlabeled input data in the test mode. 
This section explains Fuzzy-UCS with the hypertrapezoidal representation \cite{shoeleh2010handle,shoeleh2011towards}.
However, due to space constraints, the explanations provided in this section will be concise. For in-depth details, kindly refer to \cite{orriols2008fuzzy,shoeleh2010handle} and \cite{shoeleh2011towards}.

\subsection{Knowledge Representation}
An $n$-dimensional fuzzy rule $k$  is expressed by Eq. (\ref{eq: rule_repr}):
\begin{eqnarray}
\label{eq: rule_repr}
    {\rm \textbf{IF}}\; x_1 \; {\rm is} \; A_1^k \; {\rm and} \; \cdot\cdot\cdot\; {\rm and} \; x_n \; {\rm is} \; A_n^k \; {\rm \textbf{THEN}} \; c^k \; {\rm \textbf{WITH}} \; w^k ,
\end{eqnarray}
where $\bm{A}=(A_1,..., A_n)$ is a condition set, and $w^k\in[0,1]$ is a weight indicating the soundness with which the rule $k$ predicts the class $c^k$.
A fuzzy set $A_i$ is defined as $A_i=(a_i,b_i,c_i,d_i)$ by vertices of a trapezoid $a_i,b_i,c_i,d_i\in\mathbb{R}$ $(a_i\leq b_i \leq c_i\leq d_i)$ and each variable $x_i$ is represented by $A_i$ for $\forall i \in \{1,...,n\}$.

\begin{figure}[t]
\vspace{0mm}
\centerline{\includegraphics[width=0.9\linewidth]{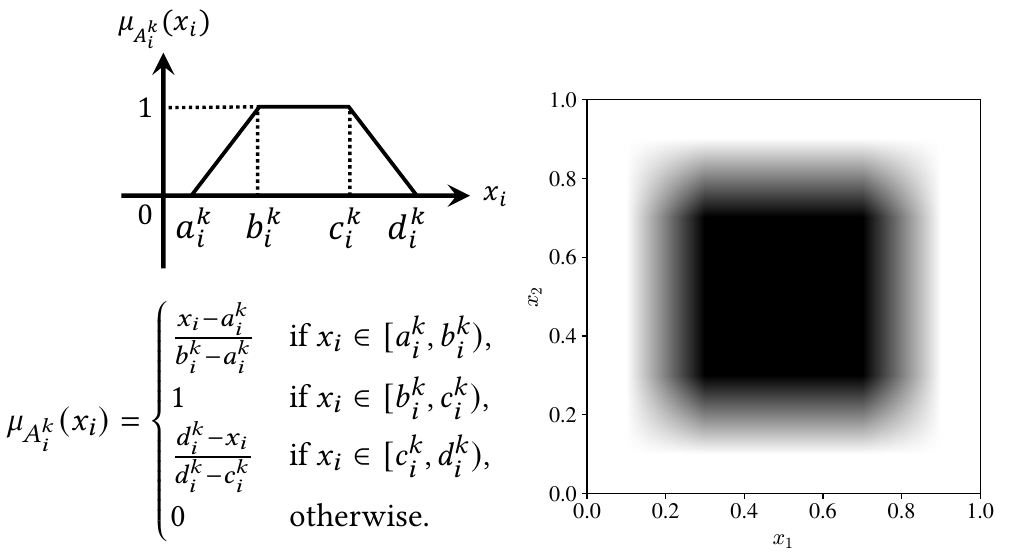}}
\vspace{0mm}
\caption{The illustration on the left depicts a fuzzy set $A_i=(a_i,b_i,c_i,d_i)$ that is characterized by a trapezoidal membership function. On the right is the illustration of the matching degree landscape of a 2-dimensional rule $k$ where $A_1^k=A_2^k=(0.1, 0.3, 0.7, 0.9)$, with white indicates $\mu_{{\bm A}^k}(\cdot)=0$, black indicates $\mu_{{\bm A}^k}(\cdot)=1$ (i.e., a \textit{certain} region), and the shades of gray indicate $\mu_{{\bm A}^k}(\cdot)\in (0,1)$ (i.e., a \textit{vague} region). }
\vspace{-4mm}
\label{fig: hypertrapezoid}
\end{figure}

The \textit{matching degree} $\mu_{\bm{A}^k}(\bm{x})\in[0,1]$ of an input vector $\bm{x}\in\mathbb{R}^n$ with a rule $k$ is computed using $\mu_{\bm{A}^k}(\bm{x})=\prod_{i=1}^n{\mu_{A^k_i}}(x_i)$,
where ${\mu_{A^k_i}}(\cdot)\in[0,1]$ is a trapezoidal-shaped membership function representing the fuzzy set $A_i^k$, as shown in Fig. \ref{fig: hypertrapezoid}. 
If $x_i$ is unknown, the system deals with missing values by considering $\mu_{{A^k_i}}(x_i)=1$.
Fig. \ref{fig: hypertrapezoid} illustrates a schematic representation of fuzzy set $A_i$ with a trapezoidal-shaped membership function and the rule-covering region of a 2-dimensional rule.

Each rule $k$ has three primary parameters: 
(i) the fitness $F^k\in(-1, 1]$, which reflects the classification accuracy of rule $k$; 
(ii) the experience $exp^k\in\mathbb{R}^+_0$, which denotes the number of times rule $k$ has been matched; 
and (iii) the numerosity $num^k\in\mathbb{N}_0$, indicating the number of encapsulated micro-rules within rule $k$. 
These parameters are continuously updated throughout the training process.

\subsection{Mechanism}
\subsubsection{Training Mode}
\label{sss: training mode Fuzzy-UCS}
At the time $t$, the system receives an input $\bm{x}$ from the environment that belongs to class $c$. Subsequently, a \textit{match set} ${\rm [M]} = \{k \in {\rm [P]} \mid \mu_{\bm{A}^k}(\bm{x}) > 0\}$ is constructed from the \textit{ruleset} [P].
After [M] is formed, the system forms a \textit{correct set} ${\rm [C]}=\{k\in{\rm [M]} \mid c^k = c\}$.
If $\sum_{k\in{\rm [C]}}{\mu_{\bm{A}^k}}(\bm{x})<1$,
the \textit{covering} operator generates a new rule $k_{\rm cov}$ such that $\mu_{\bm{A}^{k_{\rm cov}}}(\bm{x})=1$. 
This newly generated rule $k_{\rm cov}$ is then inserted into both [P] and [C].

After [C] is formed, the parameters of all rules in [M] are updated.
First, the experience is updated according to the current matching degree, as $exp^k_{t+1} = exp^k_t+\mu_{\bm{A}^k}(\bm{x})$.
Next, the fitness is updated. 
To accomplish this, each rule $k$ maintains an internally stored weight vector $\{v_1^k,...,v_m^k\}$ that is associated with the set of class labels $\{c_1,...,c_m\}$.
$m$ denotes the number of classes that can be taken, and $v_j^k$ indicates the soundness with which rule $k$ predicts class $j$ for a matched input with a matching degree of 1. 
These weights are updated by the following procedure:
\begin{enumerate}
    \item The sum of correct matchings $cm_j^k$ for each class $j$ is updated
    using $cm^k_{j_{t+1}}=cm^k_{j_t}+ \mu_{\bm{A}^k}(\bm{x})$ if $j=c$ else $cm^k_{j_t}$; 

    \item The weights $v_j^k$ for $\forall j$ is updated 
    using $v^k_{j_{t+1}}=cm^k_{j_{t+1}}/{exp^k_{t+1}}$;

    \item The fitness $F^k$ is updated 
    using $F^k_{t+1}=v^k_{{\rm max}_{t+1}} - \sum_{j\mid j\neq{\rm max}}{v^k_{j_{t+1}}}$ \cite{ishibuchi2005rule}, 
where 
the system subtracts the values of the other weights from the weight with maximum value $v^k_{\rm max}$.
%
\end{enumerate}
Finally, the highest weight, $w_c^k$, in the weight vector held by rule $k$ and its associated class label $c$ are assigned to the WITH part and the THEN part of rule $k$ in Eq. (\ref{eq: rule_repr}), respectively.

After the rules update, a steady-state GA is applied to [C].
The GA is activated when the average of the last time the GA was applied to a rule in [C] exceeds the hyperparameter $\theta_{\rm GA}$.
In this case, the two parent rules from [C] are selected through tournament selection \cite{butz2003tournament}, with the tournament size ratio determined by the hyperparameter $\tau$. 
In Fuzzy-UCS, tournament selection is carried out by the following procedure:
(i) A random sample of $\tau\times \sum_{k \in {\rm [C]} \mid F^k\geq 0}num^k$ rules is selected from  [C], excluding any rules with negative fitness;
(ii) The rule $k$ with the highest value of $(F^k)^\nu \cdot \mu_{\bm{A}^k}(\bm{x})$, where $\nu$ is a hyperparameter that penalizes the fitness, is chosen as the parent rule from the sample obtained in (i). 
The two parent rules are replicated as two child rules, with crossover and mutation applied with probabilities $\chi$ and $p_{\rm mut}$, respectively.
The two child rules are inserted into [P]. 
In order to maintain the ruleset size $N$, if the number of micro-rules in [P] exceeds this limit, the rule $k$ with relatively low powered fitness $(F^k)^\nu$ is removed preferentially.

The \textit{subsumption} operator \cite{wilson1998generalization} is employed to prevent inserting over-specific rules into the ruleset \cite{liu2020absumption}.
Subsumption is triggered after [C] is formed or after GA is executed, and is referred to as \textit{Correct Set Subsumption} and \textit{GA Subsumption}, respectively. 
Specifically, for the two rules $k_{\rm sub}$ and $k_{\rm tos}$, if 
(i) the \textit{is-more-general} operator determines that $k_{\rm sub}$ is more general than $k_{\rm tos}$ (cf. \cite{shoeleh2011towards}), 
(ii) $k_{\rm sub}$ is accurate (i.e., $F^{k_{\rm sub}}>F_0$), and 
(iii) $k_{\rm sub}$ is sufficiently experienced (i.e., $exp^{k_{\rm sub}}>\theta_{\rm sub}$), 
then $k_{\rm tos}$ is removed from [P], and $num^{k_{\rm sub}}$is updated using $num^{k_{\rm sub}} \leftarrow num^{k_{\rm sub}} + num^{k_{\rm tos}}$ to record the number of subsumed rules.

\subsubsection{Test Mode}
Given an input $\bm{x}$, all the well-updated rules within [M] cast a vote for the class they predict. 
The number of votes for each class $j$ is initially determined through the calculation expressed in ${\rm vote}_j=\sum_{k\in {\rm [M]}\mid c^k=j \, \land \, exp^k>\theta_{\rm exploit}}{v_k}$,
where $v_k=F^k\cdot\mu_{\bm{A}^k}(\bm{x})\cdot num^k$ is the number of votes cast for the class $c^k$ supported by rule $k$ and $\theta_{\rm exploit}$ is a hyperparameter.
Finally, the class with the highest number of votes is output as the inference result of the system.

\section{Proposed System: Adaptive-UCS}
\label{sec: proposed}

This section proposes a \textit{sUpervised Self-Adaptive-Classifier System} (Adaptive-UCS), a Fuzzy-UCS with a self-adaptive rule representation mechanism.
 The following subsections delineate the differences between Adaptive-UCS and Fuzzy-UCS in the following aspects: (1) rule parameters, (2) matching degree calculation, (3) covering, (4) evolutionary operators for self-adaptation, and (5) subsumption.
 
 \subsection{Rule Parameters}
 In Adaptive-UCS, a fuzzy set $A_i$ in a rule condition is specified by the center $c_i\in\mathbb{R}$ and spread $s_i\in\mathbb{R}^+$ of either a rectangular or triangular-shaped membership function, i.e., $A_i=(c_i,s_i)$. 
 
\begin{figure}[t]
\centerline{\includegraphics[width=0.9\linewidth]{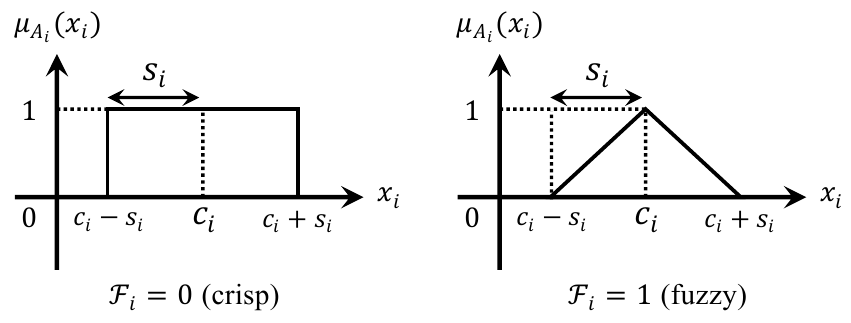}}
\vspace{0mm}
\caption{A fuzzy set $A_i=(c_i,s_i)$ in Adaptive-UCS. The membership function of $A_i$ is self-adapted based on 
the fuzzy indicator 
$\mathcal{F}_i\in\mathbb{B}$.}
\vspace{-5mm}
\label{fig: rule_repr}
\end{figure}

 As a novel rule parameter, Adaptive-UCS introduces a \textit{fuzzy indicator} $\bm{\mathcal{F}}\in\mathbb{B}^n$ that determines the membership function of the fuzzy set.
 Specifically, the fuzzy set $A_i^k$ of rule $k$ in $\mathcal{F}^k_i=0$ (i.e., crisp) is represented by a rectangular-shaped membership function with lower and upper bounds of $c_i^k-s_i^k$ and $c_i^k + s_i^k$, respectively. 
 This representation is equivalent to the \textit{center-spread hyperrectangular representation} \cite{wilson1999xcsr} in which the crisp-hyperrectangular rule condition is described by a combination of the center $\bm{c}$ and the spread $\bm{s}$.
Conversely, the fuzzy set $A_i^k$ of rule $k$ in $\mathcal{F}^k_i=1$ (i.e., fuzzy) is represented by an isosceles triangular-shaped membership function with vertices at $c_i^k-s_i^k$, $c_i^k$, and $c_i^k + s_i^k$, from left to right, respectively. 
This representation corresponds to a special form of the \textit{non-grid-oriented hypertriangular representation} \cite{orriols2011fuzzy} in which the fuzzy triangular rule condition is described by a combination of the left vertex $\bm{a}$, center vertex $\bm{b}$, and right vertex $\bm{c}$ ($\forall i: a_i\leq b_i \leq c_i$).
Fig. \ref{fig: rule_repr} illustrates a schematic representation of fuzzy set $A_i$ in Adaptive-UCS.

\subsection{Matching Degree Calculation}
In Adaptive-UCS, the calculation of the matching degree between a given input $\bm{x}$ and a rule $k$ is based on the fuzzy indicator $\bm{\mathcal{F}}^k$. 
The matching degree $\mu_{\bm{A}^k}(\bm{x};\bm{\mathcal{F}}^k)$ is computed using $\mu_{\bm{A}^k}(\bm{x};\bm{\mathcal{F}}^k)=\prod_{i=1}^n{\mu_{A^k_i}}(x_i;\mathcal{F}^k_i)$,
where ${\mu_{A^k_i}}(x_i; \mathcal{F}_i^k)$ is computed as follows:
\begin{equation}  \label{eq: proposed_membership}
{\mu_{A^k_i}}(x_i; \mathcal{F}_i^k)=
    \begin{cases}
        1      &\text{if}\;\; \mathcal{F}_i^k=0 \land x_i\in [c_i^k-s_i^k,c_i^k+s_i^k),\\
        \frac{x_i-c_i^k+s_i^k}{s_i^k}   &   \text{if}\;\; \mathcal{F}_i^k=1 \land x_i\in[c_i^k-s_i^k, c_i^k),\\
        \frac{c_i^k+s_i^k-x_i}{s_i^k} & \text{if}\;\; \mathcal{F}_i^k=1 \land x_i\in[c_i^k, c_i^k+s_i^k),\\
        0& \text{otherwise.}
    \end{cases}
\end{equation}

\begin{figure*}[t]
 \vspace{-2mm}
\captionsetup[subfigure]{justification=centering}
 \begin{minipage}[b]{0.24\linewidth}
  \centering
  \includegraphics[keepaspectratio, scale=0.27]
  {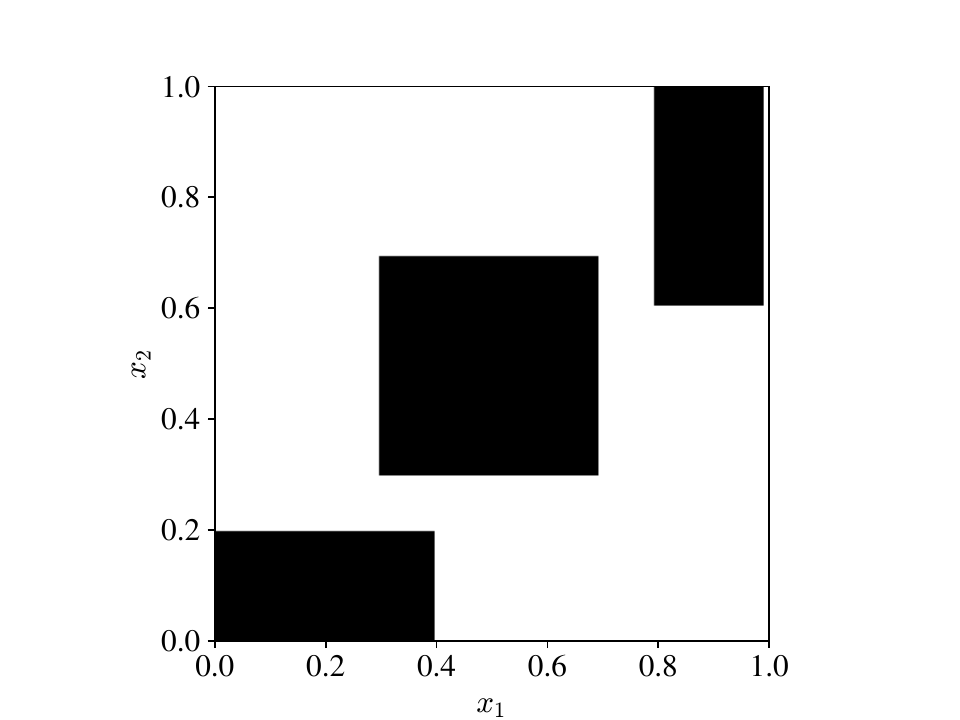}
  \vspace{-5mm}
  \subcaption{$\bm{\mathcal{F}}=(0, 0)$}\label{fig: psi_r_r}
 \end{minipage}
  \begin{minipage}[b]{0.24\linewidth}
  \centering
  \includegraphics[keepaspectratio, scale=0.27]
  {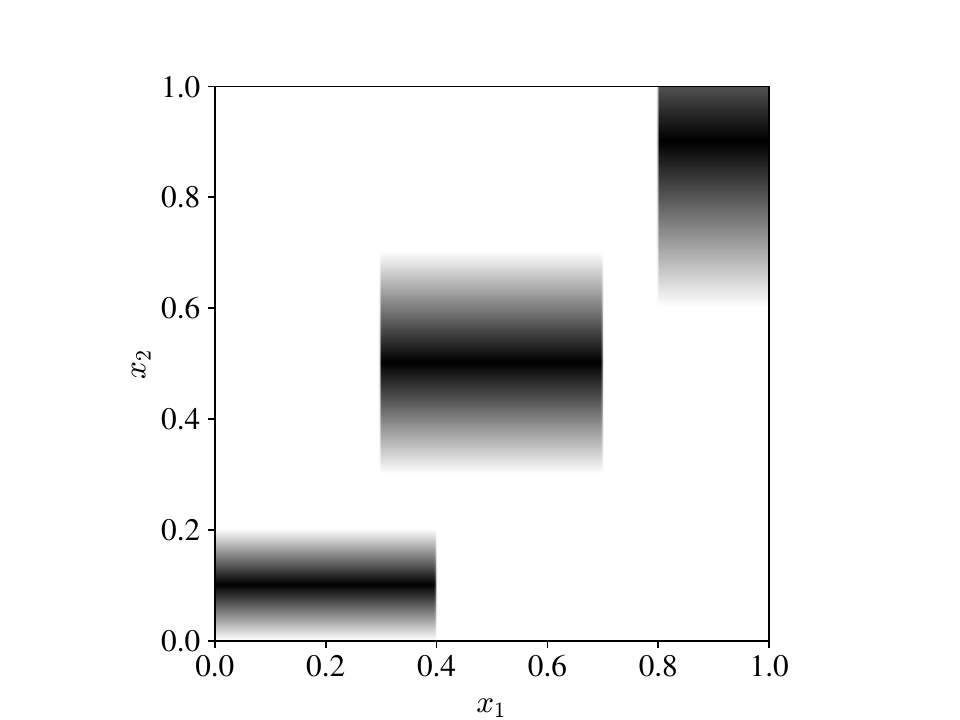}
  \vspace{-5mm}
  \subcaption{$\bm{\mathcal{F}}=(0, 1)$}\label{fig: psi_r_t}
 \end{minipage}
 \begin{minipage}[b]{0.24\linewidth}
  \centering
  \includegraphics[keepaspectratio, scale=0.27]
  {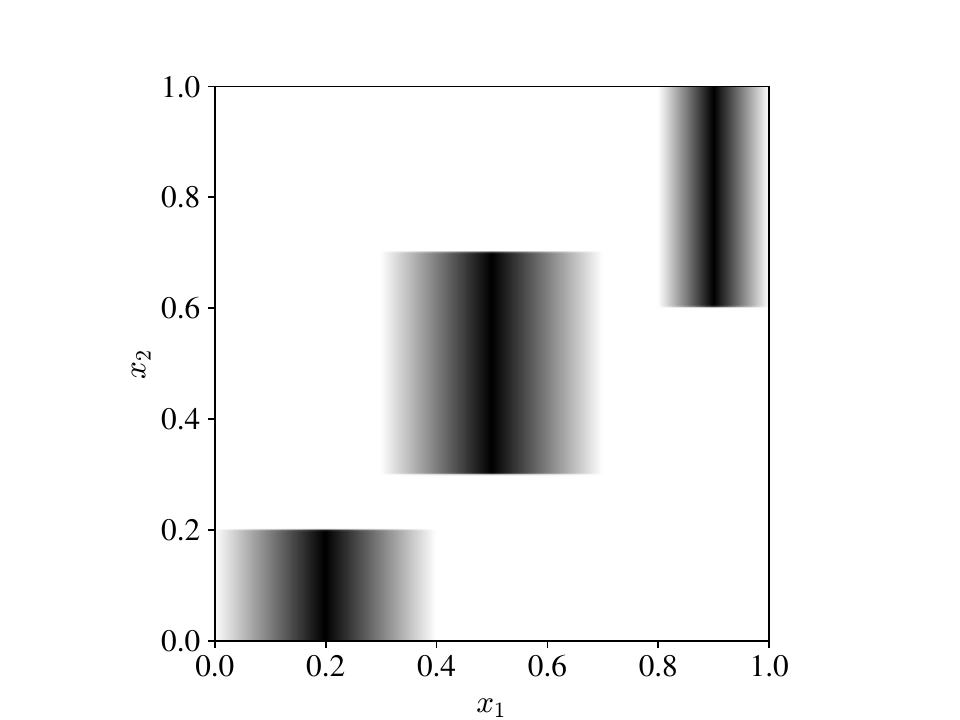}
  \vspace{-5mm}
  \subcaption{$\bm{\mathcal{F}}=(1, 0)$}\label{fig: psi_t_r}
 \end{minipage}
 \begin{minipage}[b]{0.24\linewidth}
  \centering
  \includegraphics[keepaspectratio, scale=0.27]
  {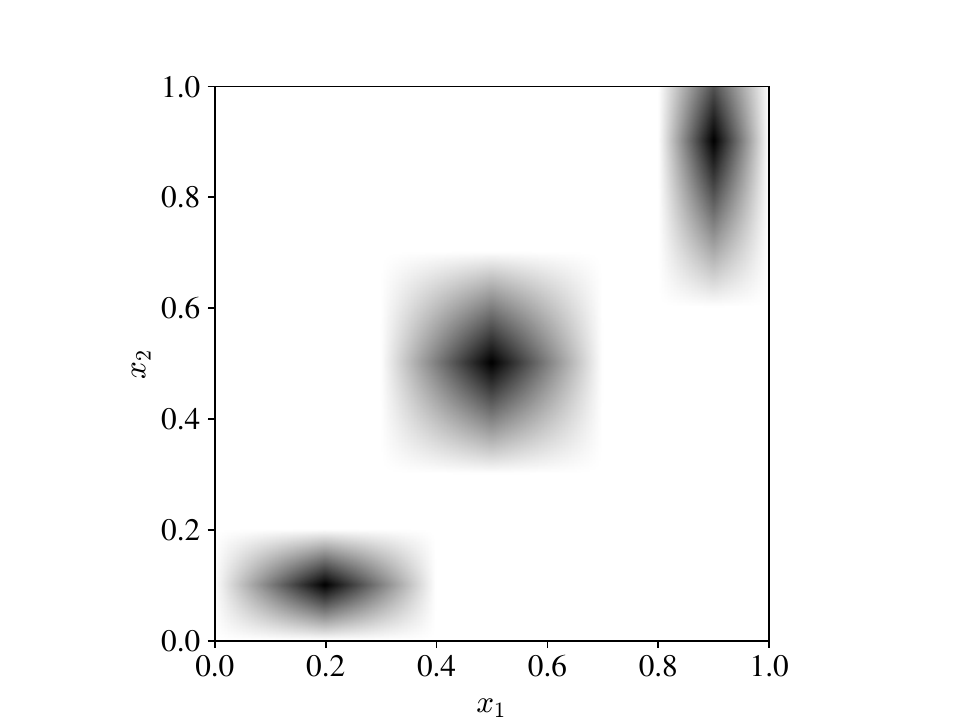}
  \vspace{-5mm}
  \subcaption{$\bm{\mathcal{F}}=(1, 1)$}\label{fig: psi_t_t}
 \end{minipage}
\vspace{-2mm}
 \caption{Matching degree landscapes of the 2D rules $k_1, k_2, k_3$ with varying fuzzy indicators $\bm{\mathcal{F}}\in\mathbb{B}^2$. Here, $k_1$, $k_2$, and $k_3$ are located in the center, lower left, and upper right, respectively, i.e.,  $A_1^{k_1}=A_2^{k_1}=(0.5,0.2)$, $A_1^{k_2}=A_2^{k_2}=(0.2,0.1)$, $ A_1^{k_3}=(0.9,0.1);A_2^{k_3}=(0.9,0.3)$.}\label{fig: proposed_rule_landscape}
 \begin{center}
\end{center}
\vspace{-2mm}
\end{figure*}

The grayscale diagrams in Fig. \ref{fig: proposed_rule_landscape} provide visual representations of the matching degree of 2-dimensional rules $k_1, k_2, k_3$ in Adaptive-UCS, where the fuzzy indicator $\bm{\mathcal{F}}$ is varied (resolution $1000 \times 1000$). 
The diagrams
demonstrate the ability of Adaptive-UCS to represent $2^n$ types of rule-covering regions with a single rule condition.  
For ease of reference, rules that satisfy $\bm{\mathcal
F}=\bm{0}$ are referred to as \textit{crisp} rules, while those that do not are referred to as \textit{fuzzy} rules.

\subsection{Covering}
 In Adaptive-UCS, the covering operator generates a new rule $k_{\rm cov}$ with fuzzy set $A_i^{k_{\rm cov}}=(c_i^{k_{\rm cov}},s_i^{k_{\rm cov}})$ for $\forall i \in \{1,...,n\}$ defined by Eq. (\ref{eq: proposed_covering}) for an $n$-dimensional input $\bm{x}$.
 \begin{equation}  \label{eq: proposed_covering}
    c_i^{k_{\rm cov}} = x_i, \quad s_i^{k_{\rm cov}} = U_{(0,r_0)},
\end{equation}
 where $r_0\in(0,1]$ is a hyperparameter specifying the maximum spread at covering and $U_{(0,r_0)}$ is a uniformly distributed random number in the range $(0,r_0)$.

The fuzzy indicator $\mathcal{F}_i^{k_{\rm cov}}$ for $\forall i \in \{1,...,n\}$ is set as in Eq. (\ref{eq: proposed_covering_psi}):
\begin{equation}  \label{eq: proposed_covering_psi}
\mathcal{F}_i^{k_{\rm cov}}=
    \begin{cases}
      0&   \text{if } U_{[0,1)} < 0.5,\\
      1 & \text{otherwise}.
    \end{cases}
\end{equation}

\subsection{Evolutionary Operators for Self-Adaptation of Rule Representations}
\subsubsection{Crossover}
 In Adaptive-UCS, crossover is applied not only to the fuzzy set $A_i=(c_i, s_i)$, but also to the fuzzy indicator $\mathcal{F}_i$, as a unique operation specific to Adaptive-UCS.

 \subsubsection{Mutation}
In Adaptive-UCS, mutation is applied to both the fuzzy set $A^k_i=(c^k_i, s^k_i)$ and the fuzzy indicator $\mathcal{F}_i^k$ of the rule $k$.

The mutation applied to the center $c_i^k$ is defined as follows:
\begin{equation} \label{eq: proposed_mutation_center}
c_i^k \leftarrow
c_i^k + U_{[-m_0, m_0)},
\end{equation}
where $m_0\in(0,1]$ signifies the maximum mutation magnitude.

The mutation applied to the spread $s_i^k$ is specified by Eq. (\ref{eq: proposed_mutation_spread}):
\begin{equation} \label{eq: proposed_mutation_spread}
s_i^k\leftarrow
\begin{cases}
s_i^k + U_{[0,m_0)} & \text{if } F^k>F_0 \text{ and no cross. has taken place,}\\
s_i^k + U_{[-m_0, m_0)}& \text{otherwise.}
\end{cases}
\end{equation}
This equation aims to enforce generalization of rule $k$ if $k$ is accurate (i.e., $F^k>F_0$) and not generated through a crossover. 
In other cases, generalization or specialization is performed. This configuration enhances the evolutionary pressure towards obtaining an accurate and general ruleset.

The mutation applied to the fuzzy indicator $\mathcal{F}_i^k$ is defined as follows:
\begin{equation}  \label{eq: proposed_mutation_psi}
\mathcal{F}_i^{k}\leftarrow
    \begin{cases}
      0 &   \text{if } \mathcal{F}_i^k=1,\\
       1 & \text{otherwise}.
    \end{cases}
\end{equation}

 \subsection{Subsumption}
 \begin{figure}[t]
\vspace{-5mm}
\centerline{\includegraphics[width=0.9\linewidth]{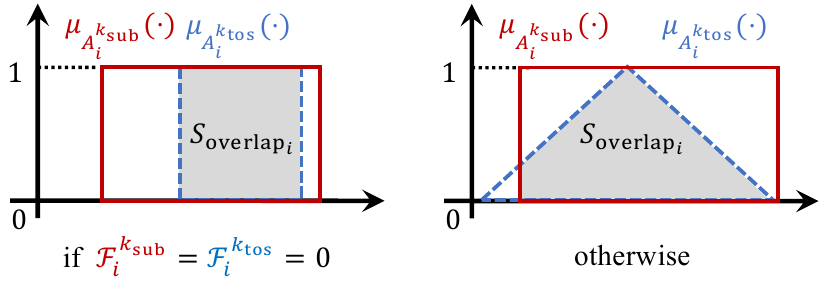}}
\vspace{-3mm}
\caption{An example of how is-more-general operator determines that $k_{\rm sub}$ is more general than $k_{\rm tos}$ in dimension $i$ with $\theta_{\rm overlap}=0.5$ in Adaptive-UCS. The gray area represents the overlap region between the two membership functions.}
\vspace{-4mm}
\label{fig: is_more_general}
\end{figure}
 \subsubsection{Is-More-General Operator}

In Adaptive-UCS, the is-more-general operator evaluates whether a rule $k_{\rm sub}$ is more general than a rule $k_{\rm tos}$. 
This is determined by the satisfaction of Eq. (\ref{eq: proposed_subsumption}) in all dimensions $i\in\{1,...,n\}$.
\begin{equation}  \label{eq: proposed_subsumption}
S_{{\rm overlap}_i} \geq
    \begin{cases}
S_i^{k_{\rm tos}} \cdot 1   & \text{if } \mathcal{F}_i^{k_{\rm sub}}=\mathcal{F}_i^{k_{\rm tos}}=0,\\
S_i^{k_{\rm tos}} \cdot \theta_{\rm overlap}  &\text{otherwise}.
    \end{cases}
\end{equation}

In Eq. (\ref{eq: proposed_subsumption}), $S_{{\rm overlap}_i}$ represents the area of the overlap region between ${\mu_{A^{k_{\rm sub}}_i}}(\cdot)$ and ${\mu_{A^{k_{\rm tos}}_i}}(\cdot)$.
$S_i^{k_{\rm tos}}$ is the total area of ${\mu_{A^{k_{\rm tos}}_i}}(\cdot)$. 
$\theta_{\rm overlap}\in(0,1]$ is the scaling parameter.
The significance of Eq. (\ref{eq: proposed_subsumption}) can be expressed as follows. 
When both $A^{k_{\rm sub}}_i$ and $A^{k_{\rm tos}}_i$ are rectangular-shaped membership functions with \textit{certain} bounds, Eq. (\ref{eq: proposed_subsumption}) is satisfied only if ${\mu_{A^{k_{\rm sub}}_i}}(\cdot)$ completely encompasses ${\mu_{A^{k_{\rm tos}}_i}}(\cdot)$. 
On the other hand, if either $A^{k_{\rm sub}}_i$ or $A^{k_{\rm tos}}_i$, or both, are triangular-shaped membership functions with \textit{vague} bounds, then Eq. (\ref{eq: proposed_subsumption}) is satisfied if ${\mu_{A^{k_{\rm sub}}_i}}(\cdot)$ covers a region proportional to $\theta_{\rm overlap}$ times that of ${\mu_{A^{k_{\rm tos}}_i}}(\cdot)$\footnote{The upper and lower conditionals specified in Eq. (\ref{eq: proposed_subsumption}) correspond to the is-more-general operators in the crisp-hyperrectangular \cite{wilson1999xcsr} and fuzzy-hypertrapezoidal \cite{shoeleh2010handle,shoeleh2011towards} representations, respectively.}.
Fig. \ref{fig: is_more_general} schematically depicts the scenario under which Eq. (\ref{eq: proposed_subsumption}) holds true for dimension $i$ when $\theta_{\rm overlap}=0.5$.

 \subsubsection{Subsumption with Merge Mechanism}
\begin{figure}[t]
 \vspace{-5mm}
\centerline{\includegraphics[width=0.95\linewidth]{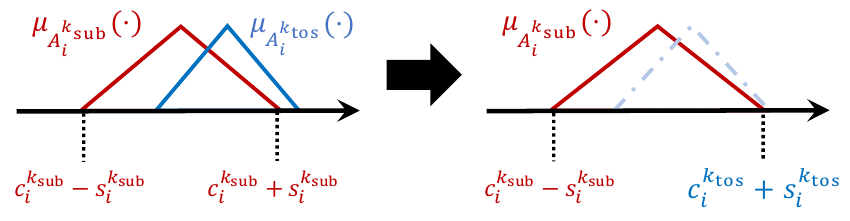}}
\vspace{-3mm}
\caption{An example of how merge mechanism works in Adaptive-UCS.}
\vspace{-5mm}
\label{fig: merge}
\end{figure}

In Adaptive-UCS, rules merge when specific criteria are satisfied during the subsumption process. 
This merge mechanism is an adapted version of the merge mechanism originally proposed for the fuzzy-hypertrapezoidal representation \cite{shoeleh2011towards} in Adaptive-UCS. Specifically, just like in the Fuzzy-UCS approach, a rule $k_{\rm sub}$ that is more general, accurate, and well-experienced than rule $k_{\rm tos}$ is first identified (cf. Sect. \ref{sss: training mode Fuzzy-UCS}). If $k_{\rm tos}$ is also accurate and well-experienced (i.e., $F^{k_{\rm tos}}>F_0 \land exp^{k_{\rm tos}}>\theta_{\rm sub}$), and there exists a dimension $i$ such that $\mathcal{F}_i^{k_{\rm sub}} = \mathcal{F}_i^{k_{\rm tos}} = 1$, then $A_i^{k_{\rm sub}}=(c_i^{k_{\rm sub}},s_i^{k_{\rm sub}})$ is updated according to Eq. (\ref{eq: proposed_merge}):
  \begin{equation}  \label{eq: proposed_merge}
    c_i^{k_{\rm sub}} \leftarrow (u_i+l_i)/2, \quad s_i^{k_{\rm sub}} \leftarrow (u_i-l_i)/2,
\end{equation}
where $u_i = \max \{c_i^{k_{\rm sub}} + s_i^{k_{\rm sub}}, c_i^{k_{\rm tos}} + s_i^{k_{\rm tos}}\}$ and $l_i = \min \{c_i^{k_{\rm sub}} - s_i^{k_{\rm sub}}, c_i^{k_{\rm tos}} - s_i^{k_{\rm tos}}\}$.
Fig. \ref{fig: merge} illustrates the schematic representation of how $k_{\rm sub}$ merges and elevates the generality of $k_{\rm tos}$ in dimension \textit{i}.
 
Note 
that the implementation of the merge mechanism is restricted to instances where both $A^{k_{\rm sub}}_i$ and $A^{k_{\rm tos}}_i$ are represented by triangular-shaped membership functions with \textit{vague} bounds, i.e., $\mathcal{F}_i^{k_{\rm sub}} = \mathcal{F}_i^{k_{\rm tos}} = 1$. 
This restriction has been imposed to alleviate the risk of \textit{over-generalization} of rules, a well-established concern in the context of crisp-hyperrectangular representation \cite{wagner2022mechanisms, shiraishi2022absumption}, and minimize its detrimental impact on the system's performance.

\section{Experiment 1: Benchmark Problems}
\label{sec: experiment 1}
\subsection{Problem Description}
\subsubsection{Checkerboard Problem}
The \textit{checkerboard} (CB) problem \cite{stone2003real} is a commonly used benchmark problem in the evaluation of real-valued LCSs. 
The CB problem involves a randomly generated two-dimensional real-valued input vector, $\bm{x}\in [0,1)^2$, that must be classified into one of two answer classes, $\{0, 1\}$, based on its presence in either a black or white region.
Given that the checkerboard pattern can be divided into simple rectangles, and the rule-covering regions of all three systems are based on rectangles, these systems are expected to yield high classification performance. In this paper, the state space of the CB problem was divided into five parts.
\subsubsection{Rotated Checkerboard Problem}
The \textit{rotated checkerboard} (RCB) problem \cite{shiraishi2022absumption} is a variation of the CB problem that has been rotated by 45 degrees. Unlike the CB problem, the RCB problem is comprised of diagonal class boundaries, which can pose a significant challenge for UCS, which utilize rectangular rules to approximate the class boundaries. This may result in the requirement of a large number of specific rectangular rules, making it difficult to generalize the ruleset effectively. 
\subsubsection{Noisy Checkerboard Problem}

An \textit{noisy checkerboard} (NCB) problem is the CB problem with uncertainty. During both training and test modes, the environment (same as CB) transmits an input vector $\bm{x}$ that belongs to the correct answer class $c$ to the UCSs.
However, only during training mode, before being received by the UCSs, $\bm{x}$ is subjected to Gaussian noise $\mathcal{N}(0,\sigma^2)$ which is introduced as $\forall i\in\{1,2\}:x_i\leftarrow x_i + \mathcal{N}(0,\sigma^2)$. 
Given the robustness of fuzzy rules against uncertainty, it is expected that both Fuzzy-UCS and Adaptive-UCS will perform better in this problem scenario.
In this paper, $\sigma$ is set to 0.05.

\subsection{Experimental Setup}
\label{ss: experiment1 setup}
For all of the considered problems, the hyperparameters for UCS, Fuzzy-UCS, and Adaptive-UCS were set to the values established in prior research \cite{orriols2008fuzzy,shoeleh2011towards}. 
Specifically, the following configurations were implemented: (i) UCS was set as follows: $N=6400$, $acc_{0}=0.99$, $\beta=0.2$, $\nu=10$, $\chi=0.8$, $p_{\rm mut}=0.04$, $\delta=0.1$, $m_0=0.1$, $r_0=0.2$, $\{\theta_{\rm GA},\theta_{\rm del},\theta_{\rm sub}\}=50$, $\tau=0.4$, $doCorrectSetSubsumption=yes$, $doGASubsumption=yes$; (ii) Fuzzy-UCS was set as in UCS, with the exceptions of $acc_0=\text{N/A}, \beta=\text{N/A}$, $F_{0}=0.99$, $\theta_{\rm exploit}=10$, $\theta_{\rm overlap}=0.8$; (iii) Adaptive-UCS was set as in Fuzzy-UCS, except $\theta_{\rm overlap}=0.5$.
  For the NCB problem, which is characterized by a noisy environment, we modified $\nu=1$ for all UCSs as per \cite{urbanowicz2015exstracs}, and as recommended in \cite{urbanowicz2017introduction}, $acc_0=0.95$ for UCS, $F_0=0.95$ for Fuzzy-UCS and Adaptive-UCS.
In UCS, the \textit{unordered bound hyperrectangular representation} \cite{stone2003real} were utilized for the rule conditions, and the rule merge mechanism was activated during the subsumption execution of Fuzzy-UCS and Adaptive-UCS.
The uniform crossover was used for all UCSs in the GA.
The number of alternating training/test steps was set to 200,000.
The evaluation of the systems is based on two criteria: average classification accuracy and average ruleset size. To gauge the performance, 30 independent experiments were conducted, each using a different random seed, and the evaluation values (i.e., classification accuracy and ruleset size) were recorded at two stages of the test mode. The first stage, (i), evaluated the overall performance of the system by computing the average evaluation values from the 1-200,000th test. The second stage, (ii), evaluated the convergence performance of the system by computing the average evaluation values from the 190,001-200,000th test. The evaluation values were analyzed for \textit{homoscedasticity} through the application of \textit{Levene's test}. If \textit{homoscedasticity} was positive, \textit{One-Way ANOVA} and \textit{Tukey-HSD} post-hoc test were conducted in pairs. If \textit{homoscedasticity} was negative, \textit{Welch-ANOVA} and \textit{Games-Howell} post-hoc test were conducted in combination.

\subsection{Results}
\begin{figure*}[h]
 \begin{minipage}[b]{0.32\linewidth}
  \centering
  \includegraphics[keepaspectratio, scale=0.35]
  {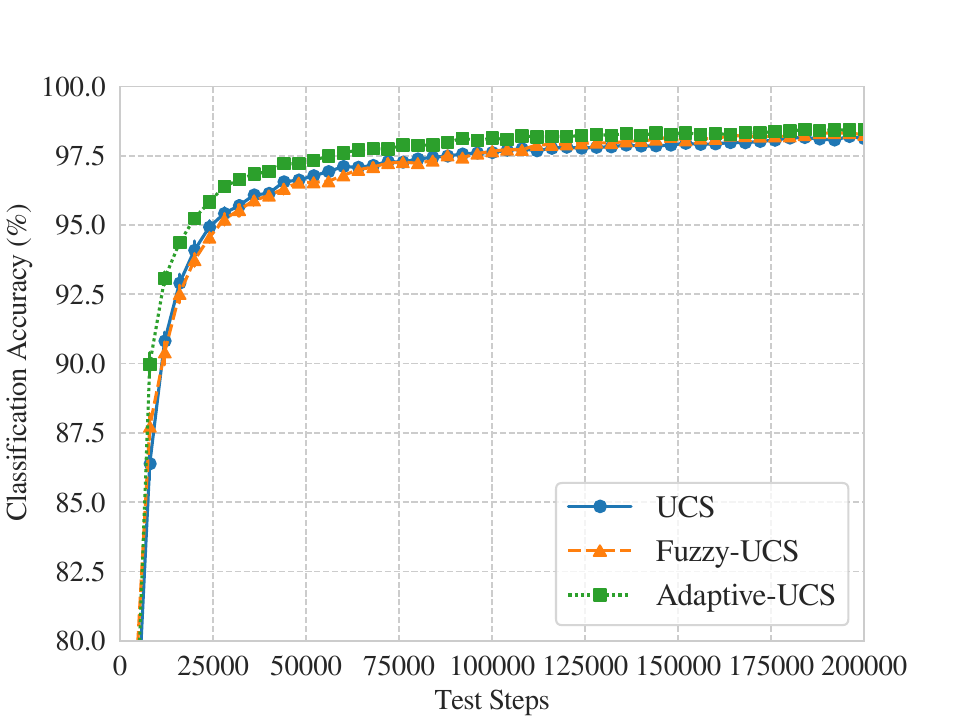}
  \vspace{-5mm}
  \subcaption{Checkerboard (CB)}\label{fig: cb_result}
 \end{minipage}
 \begin{minipage}[b]{0.32\linewidth}
  \centering
  \includegraphics[keepaspectratio, scale=0.35]
  {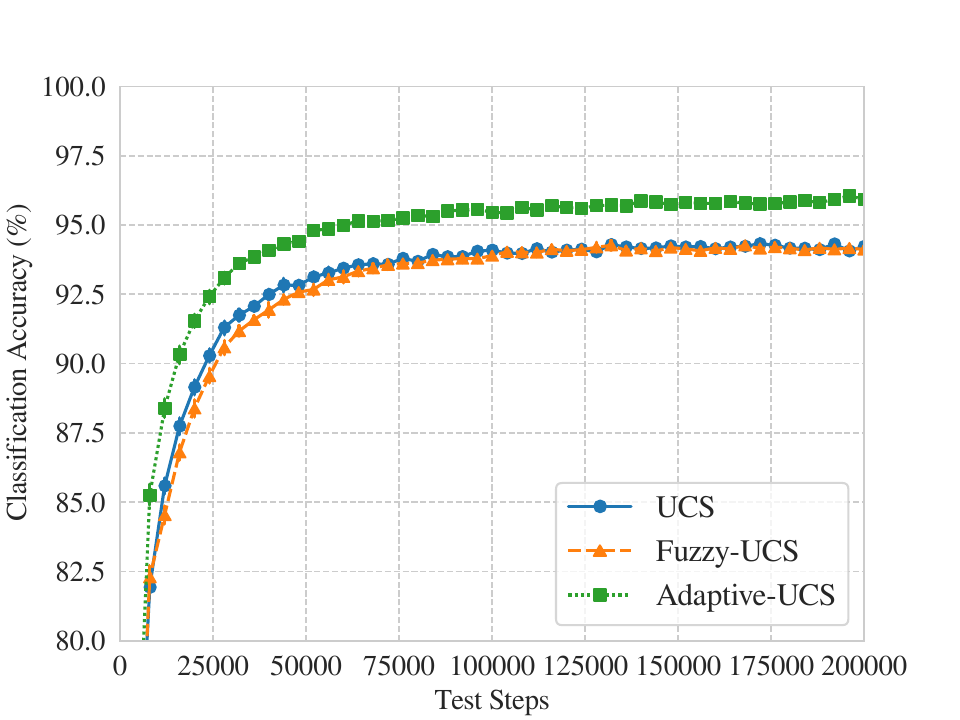}
  \vspace{-5mm}
  \subcaption{Rotated Checkerboard (RCB)}\label{fig: rcb_result}
 \end{minipage}
 \begin{minipage}[b]{0.32\linewidth}
  \centering
  \includegraphics[keepaspectratio, scale=0.35]
  {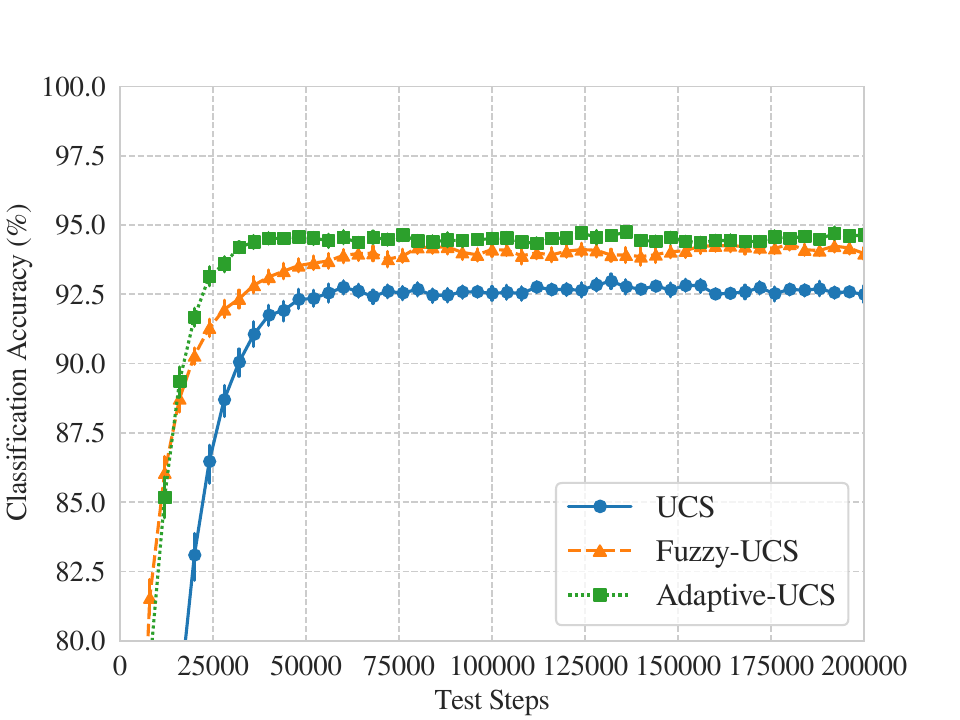}
  \vspace{-5mm}
  \subcaption{Noisy Checkerboard (NCB)}\label{fig: ncb_result}
 \end{minipage}
 \captionsetup[figure]{justification=centering}
\vspace{-2.5mm}
 \caption{UCS vs. Fuzzy-UCS vs. Adaptive-UCS on all problems in Experiment 1.}\label{fig: experiment 1}
 \begin{center}
\end{center}
\end{figure*}

Table \ref{tb: result_exp1} showcases the average evaluation values and average rank order of each evaluation value for each system across all problems.
Fig. \ref{fig: experiment 1} depicts the moving average of the classification accuracy for all systems. The horizontal axis represents the number of test steps, while the vertical axis displays the average classification accuracy computed over 30 trials. The error bars in each graph indicate the 95\% confidence interval.

As depicted in Table \ref{tb: result_exp1}, the results of the proposed Adaptive-UCS in all problems indicate that there are no peach cells, i.e., Adaptive-UCS does not belong to the worst group. Furthermore, for the CB and RCB problems, all cells are depicted in green, i.e., Adaptive-UCS belongs to the best group. This result underscores the efficacy of Adaptive-UCS across benchmark problems of varying characteristics. Additionally, as indicated by Table \ref{tb: result_exp1} and Figs. \ref{fig: rcb_result} and \ref{fig: ncb_result}, Adaptive-UCS demonstrated a remarkable enhancement in classification accuracy in comparison to either or both the conventional UCS and Fuzzy-UCS when applied to the RCB and NCB problems.

\begin{table}[t]
\begin{center}

\caption{Summary of results from Experiment 1, displaying average overall and convergence classification accuracy and ruleset size across 30 independent trials. Parentheses denote groups with statistically significant differences, with a $p$-value $<\alpha=0.05$. The rank order among all systems is indicated, with the best value denoted as ``1'' and the worst as ``2'' or ``3''. Bold green-shaded values represent the best ranks, while peach-shaded values represent the worst ranks.
\label{tb: result_exp1}}
\small
\vspace{-3mm}
\scalebox{0.9}{

\begin{tabular}{l|l|c|c|c|c}
\bhline{1pt}
\multirow{2}{*}{\textsc{Problem}} & \multirow{2}{*}{\textsc{System}} &\multicolumn{2}{c|}{\multirow{1}{*}{\textsc{Entire 200,000 Tests}}} & \multicolumn{2}{c}{\multirow{1}{*}{\textsc{Last 10,000 Tests}}}\\
& & Acc. (\%)& $|[{\rm P}]|$& Acc. (\%)& $|[{\rm P}]|$
\\
\bhline{1pt}
\multirow{3}{*}{CB}&Adaptive-UCS
& \cellcolor[rgb]{0.925, 0.957, 0.831}$\textbf{97.08}$ (1)
& \cellcolor[rgb]{0.925, 0.957, 0.831}$\textbf{2525}$ (1)
& \cellcolor[rgb]{0.925, 0.957, 0.831}$\textbf{98.34}$ (1)
& \cellcolor[rgb]{0.925, 0.957, 0.831}$\textbf{2651}$ (1)\\
\cdashline{2-6}
&Fuzzy-UCS
& \cellcolor[rgb]{0.980,0.910,0.922}96.56 (2)
& \cellcolor[rgb]{0.980,0.910,0.922}2936 (2)
& \cellcolor[rgb]{0.925, 0.957, 0.831}$\textbf{98.24}$ (1)
& \cellcolor[rgb]{0.980,0.910,0.922}$3416$ (3)\\
&UCS
& \cellcolor[rgb]{0.980,0.910,0.922}96.48 (2)
& \cellcolor[rgb]{0.980,0.910,0.922}2955 (2)
& \cellcolor[rgb]{0.925, 0.957, 0.831}$\textbf{98.15}$ (1)
& $2962$ (2)\\
\hline
\multirow{3}{*}{RCB}&Adaptive-UCS
& \cellcolor[rgb]{0.925, 0.957, 0.831}$\textbf{94.32}$ (1)
& \cellcolor[rgb]{0.925, 0.957, 0.831}$\textbf{3457}$ (1)
& \cellcolor[rgb]{0.925, 0.957, 0.831}$\textbf{95.77}$ (1)
& \cellcolor[rgb]{0.925, 0.957, 0.831}$\textbf{3594}$ (1)\\
\cdashline{2-6}
&Fuzzy-UCS
& \cellcolor[rgb]{0.980,0.910,0.922}92.42 (3)
& \cellcolor[rgb]{0.980,0.910,0.922}3801 (3)
& \cellcolor[rgb]{0.980,0.910,0.922}93.93 (2)
& \cellcolor[rgb]{0.980,0.910,0.922}4219 (3)\\
&UCS
& 92.63 (2)
& 3786 (2)
& \cellcolor[rgb]{0.980,0.910,0.922}94.36 (2)
& 3966 (2)\\
\hline
\multirow{3}{*}{NCB}&Adaptive-UCS
& \cellcolor[rgb]{0.925, 0.957, 0.831}$\textbf{93.23}$ (1)
& $4601$ (2)
& \cellcolor[rgb]{0.925, 0.957, 0.831}$\textbf{94.41}$ (1)
& $4754$ (2)\\
\cdashline{2-6}
&Fuzzy-UCS
& $92.76$ (2)
& \cellcolor[rgb]{0.980,0.910,0.922}$4688 $ (3)
& \cellcolor[rgb]{0.925, 0.957, 0.831}$\textbf{93.79} $  (1)
& \cellcolor[rgb]{0.980,0.910,0.922}$4840$ (3)\\
&UCS
& \cellcolor[rgb]{0.980,0.910,0.922}$90.27 $ (3)
& \cellcolor[rgb]{0.925, 0.957, 0.831}$\textbf{4538} $ (1)
& \cellcolor[rgb]{0.980,0.910,0.922}$92.66 $ (2)
& \cellcolor[rgb]{0.925, 0.957, 0.831}$\textbf{4668}$ (1)\\
\bhline{1pt}
\multirow{3}{*}{\textit{Avg. Rank}}&Adaptive-UCS
&\textbf{\textit{1}}
&\textbf{\textit{1.3}}
&\textbf{\textit{1}}
&\textbf{\textit{1.3}}\\
\cdashline{2-6}
&Fuzzy-UCS
&\textit{2.3}
&\textit{2.7}
&\textit{1.3}
&\textit{3}\\
&UCS
&\textit{2.3}
&\textit{1.7}
&\textit{1.7}
&\textit{1.7}\\

\bhline{1pt}
\end{tabular}
}
\end{center}
\vspace{-4mm}
\end{table}

\begin{figure*}[h]
\captionsetup[subfigure]{justification=centering}
\vspace{-0.72cm}
 \begin{minipage}[b]{0.24\linewidth}
  \centering
  \includegraphics[keepaspectratio, scale=0.27]
  {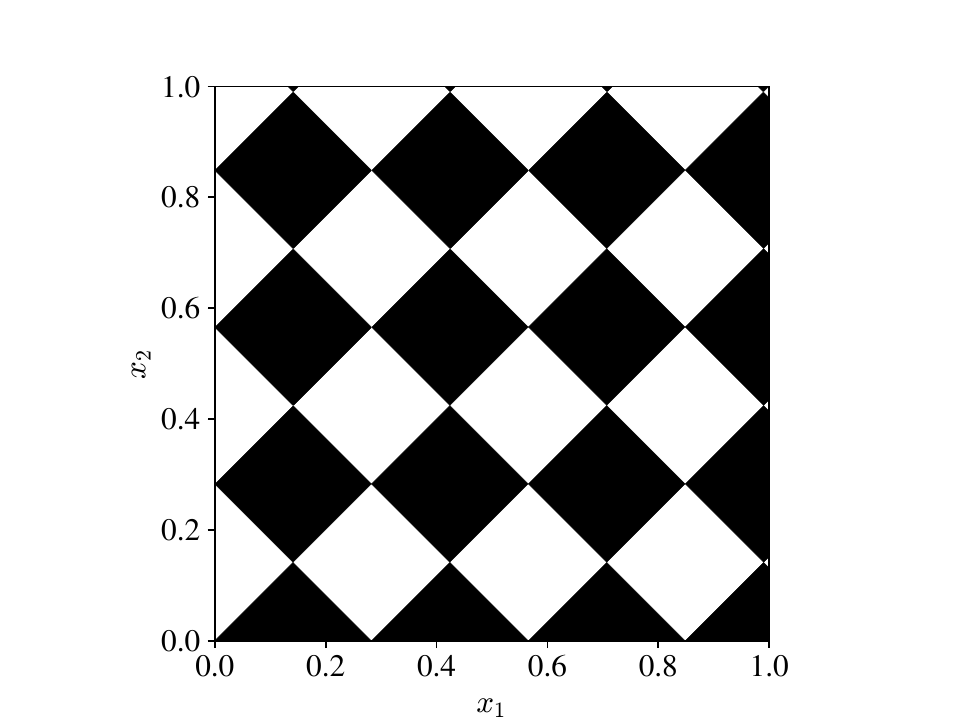}
  \vspace{-5mm}
  \subcaption{Original Data}\label{fig: rcb_landscape}
 \end{minipage}
  \begin{minipage}[b]{0.24\linewidth}
  \centering
  \includegraphics[keepaspectratio, scale=0.27]
  {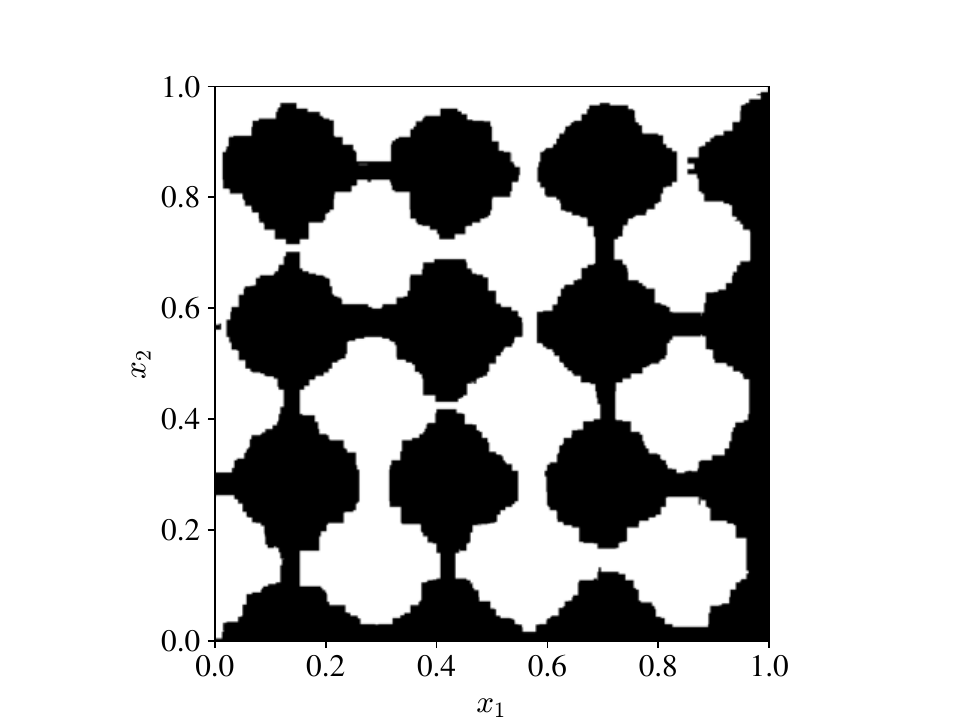}
  \vspace{-5mm}
  \subcaption{UCS (Hyperrectangles)}\label{fig: rcb_ucs_landscape}
 \end{minipage}
 \begin{minipage}[b]{0.24\linewidth}
  \centering
  \includegraphics[keepaspectratio, scale=0.27]
  {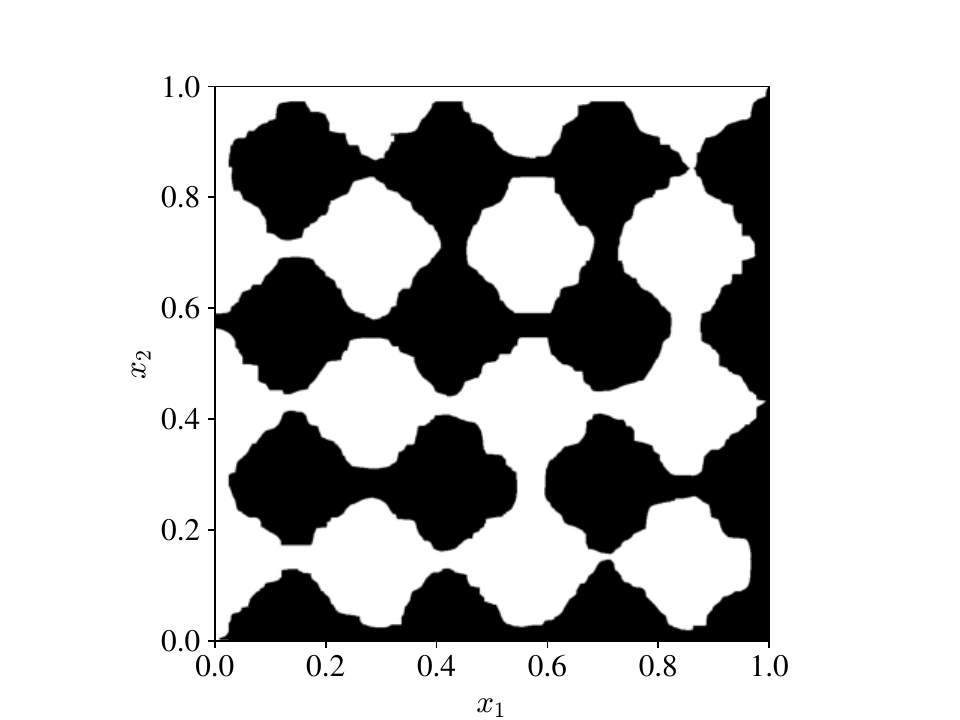}
  \vspace{-5mm}
  \subcaption{Fuzzy-UCS (Hypertrapezoids)}\label{fig: rcb_fuzzyucs_landscape}
 \end{minipage}
 \begin{minipage}[b]{0.24\linewidth}
  \centering
  \includegraphics[keepaspectratio, scale=0.27]
  {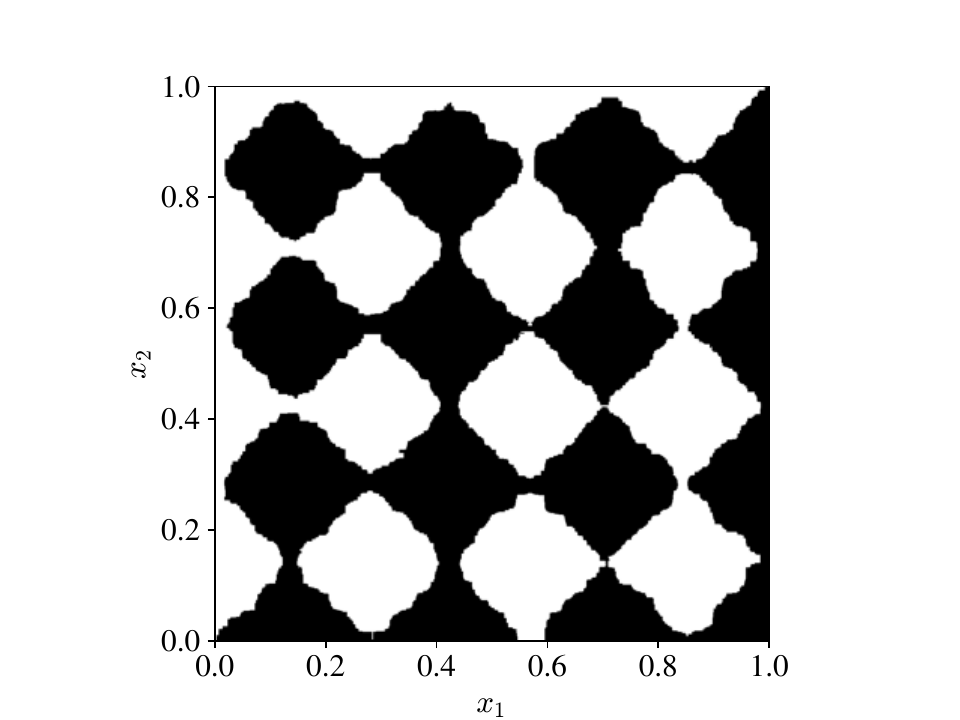}
  \vspace{-5mm}
  \subcaption{Adaptive-UCS}\label{fig: rcb_adaptiveucs_landscape}
 \end{minipage}
 \captionsetup[figure]{justification=centering}
\vspace{-2.5mm}
 \caption{(a) is the RCB problem, and (b)-(d) show landscapes of inference classes output by the system after training. In the RCB problem, the system that belongs to the best group is Adaptive-UCS (cf. Table \ref{tb: result_exp1}).}\label{fig: rcb_all_landscape}
 \begin{center}
\end{center}
\end{figure*}

\begin{figure*}[h]
\captionsetup[subfigure]{justification=centering}
\vspace{-0.72cm}
 \begin{minipage}[b]{0.24\linewidth}
  \centering
  \includegraphics[keepaspectratio, scale=0.27]
  {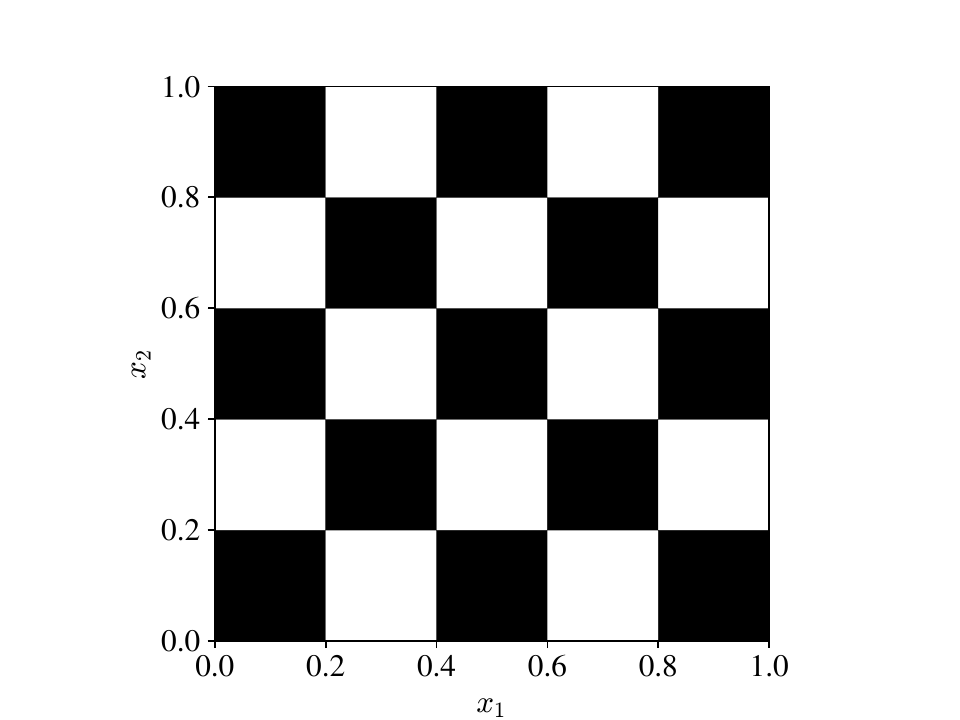}
  \vspace{-5mm}
  \subcaption{Original Data}\label{fig: cb_landscape}
 \end{minipage}
  \begin{minipage}[b]{0.24\linewidth}
  \centering
  \includegraphics[keepaspectratio, scale=0.27]
  {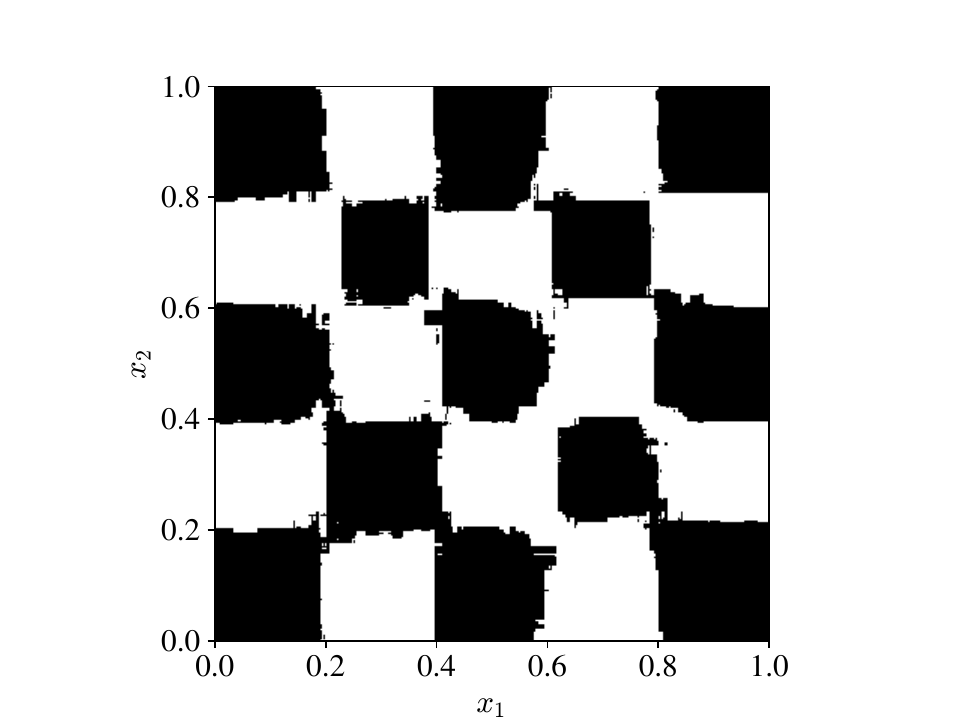}
  \vspace{-5mm}
  \subcaption{UCS (Hyperrectangles)}\label{fig: ncb_ucs_landscape}
 \end{minipage}
 \begin{minipage}[b]{0.24\linewidth}
  \centering
  \includegraphics[keepaspectratio, scale=0.27]
  {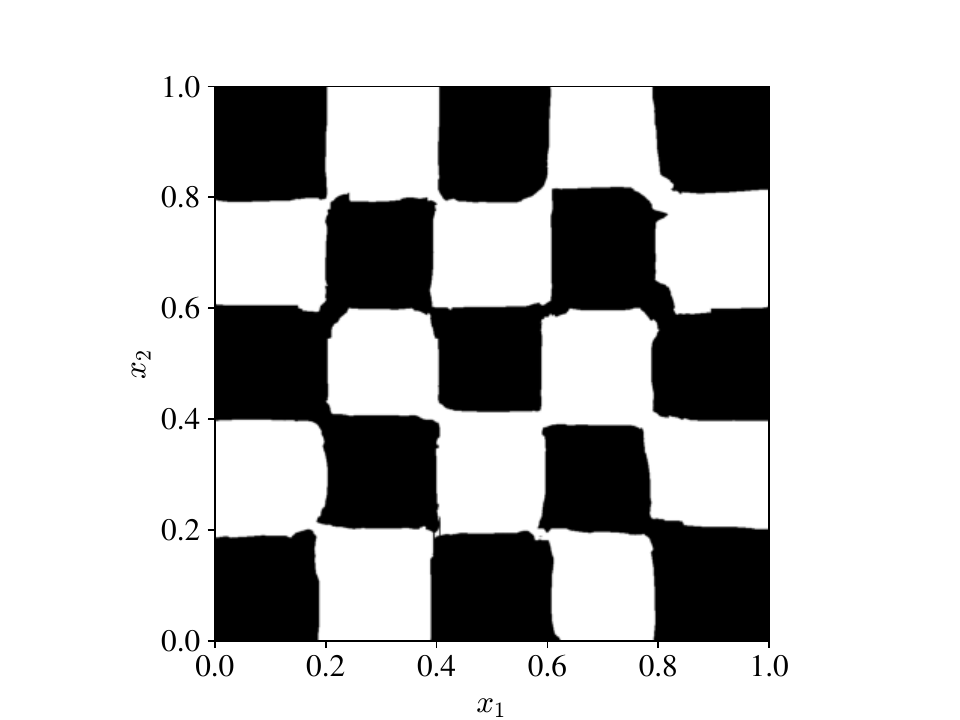}
  \vspace{-5mm}
  \subcaption{Fuzzy-UCS (Hypertrapezoids)}\label{fig: ncb_fuzzyucs_landscape}
 \end{minipage}
 \begin{minipage}[b]{0.24\linewidth}
  \centering
  \includegraphics[keepaspectratio, scale=0.27]
  {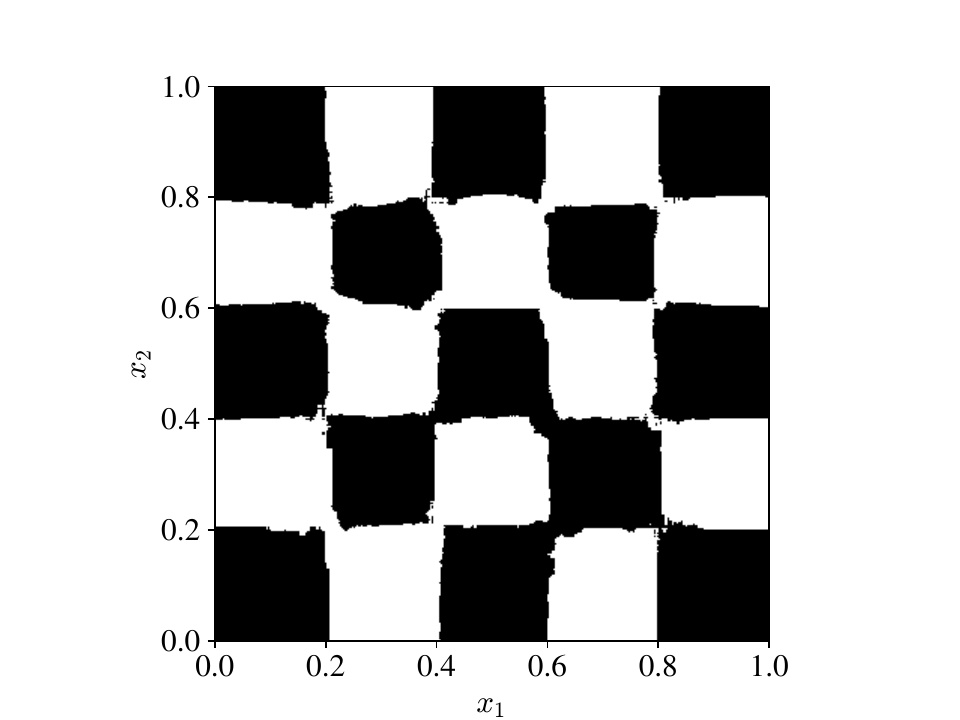}
  \vspace{-5mm}
  \subcaption{Adaptive-UCS}\label{fig: ncb_adaptiveucs_landscape}
 \end{minipage}
 \captionsetup[figure]{justification=centering}
\vspace{-2.5mm}
 \caption{(a) is the NCB problem, and (b)-(d) show landscapes of inference classes output by the system after training. In the NCB problem, the systems that belong to the best group are Fuzzy-UCS and Adaptive-UCS (cf. Table \ref{tb: result_exp1}).}\label{fig: ncb_all_landscape}
 \vspace{-5.5mm}
 \begin{center}
\end{center}
\end{figure*}

\begin{table*}[t]
\begin{center}
\caption{Summary of results from Experiment 2, displaying average classification accuracy across 30 independent trials. Numbers in parentheses, bold green-shaded values, and peach-shaded values are to be interpreted as for Table \ref{tb: result_exp1}. The left columns describe: the name of the dataset (Name), the number of instances ($\#$\textsc{Inst.}), the total number of features ($\#$\textsc{Fea.}), the number of classes ($\#$\textsc{Cl.}), the percentage of missing data attributes ($\%$\textsc{Mis.}), and the source of the dataset (Ref.). The symbol $\dag$ ($\S$) represents statistically significant differences in performance between UCS (Fuzzy-UCS) and Adaptive-UCS (denoted as Ours), i.e., that for the $p$-value of \textit{Wilcoxon signed-rank test} hold $p<\alpha=0.05$. The color-coded symbols (red and blue) signify the superiority or inferiority of Adaptive-UCS relative to UCS or Fuzzy-UCS, respectively.}
\vspace{-2mm}
\label{tb:result}
\normalsize
\scalebox{0.8475}{
\begin{tabular}{r  c c c c c|c|c|r|c|c|r}
\bhline{1pt}
\multicolumn{6}{l|}{\textsc{20 Real-World Datasets}} &\multicolumn{3}{c|}{\textsc{Training Accuracy (\%)}} & \multicolumn{3}{c}{\textsc{Test Accuracy (\%)}}\\
Name  & $\#$\textsc{Inst.}  & $\#$\textsc{Fea.} &$\#$\textsc{Cl.} & $\%$\textsc{Mis.} & Ref. & UCS&Fuzzy-UCS& \multicolumn{1}{c|}{Ours}&UCS&Fuzzy-UCS&\multicolumn{1}{c}{Ours}
\\
\bhline{1pt}
Balance scale
& 625
& 4
& 3
& 0
& \cite{Dua:2019}
&  \cellcolor[rgb]{0.980,0.910,0.922}87.40 (3)
& 88.50 (2)
&  \cellcolor[rgb]{0.925, 0.957, 0.831}{\red{$^{\dag\S}$}}\textbf{90.36} (1)
&  \cellcolor[rgb]{0.925, 0.957, 0.831}\textbf{86.40} (1)
& \cellcolor[rgb]{0.925, 0.957, 0.831}\textbf{87.37} (1)
&  \cellcolor[rgb]{0.925, 0.957, 0.831}{\red{$^{\dag\S}$}}\textbf{89.30} (1)\\
Column 3C weka
& 310
& 6
& 3
& 0
& \cite{Dua:2019}
&  \cellcolor[rgb]{0.980,0.910,0.922}73.55 (2)
&  \cellcolor[rgb]{0.980,0.910,0.922}73.41 (2)
& \cellcolor[rgb]{0.925, 0.957, 0.831} {\red{$^{\dag\S}$}}\textbf{82.10} (1)
&  \cellcolor[rgb]{0.980,0.910,0.922}71.83 (2)
&  \cellcolor[rgb]{0.980,0.910,0.922}68.92 (2)
& \cellcolor[rgb]{0.925, 0.957, 0.831} {\red{$^{\dag\S}$}}\textbf{77.53} (1)\\
Diabetes
& 768
& 8
& 2
& 0
& \cite{smith1988using}
& 76.66 (2)
&  \cellcolor[rgb]{0.980,0.910,0.922}75.49 (3)
&  \cellcolor[rgb]{0.925, 0.957, 0.831}{\red{$^{\dag\S}$}}\textbf{78.43} (1)
& \cellcolor[rgb]{0.925, 0.957, 0.831}\textbf{72.06} (1)
&  \cellcolor[rgb]{0.925, 0.957, 0.831}\textbf{72.54} (1)
&  \cellcolor[rgb]{0.925, 0.957, 0.831}{\red{$^{\dag\S}$}}\textbf{74.47} (1)\\
Ecoli
& 336
& 7
& 8
& 0
& \cite{Dua:2019}
&  \cellcolor[rgb]{0.980,0.910,0.922}77.30 (2)
&  \cellcolor[rgb]{0.980,0.910,0.922}83.19 (2)
&  \cellcolor[rgb]{0.925, 0.957, 0.831}{\red{$^{\dag\S}$}}\textbf{87.96} (1)
&  \cellcolor[rgb]{0.980,0.910,0.922}73.03 (3)
&  78.09 (2)
&  \cellcolor[rgb]{0.925, 0.957, 0.831}{\red{$^{\dag\S}$}}\textbf{83.13} (1)\\
Fruit
& 898
& 34
& 7
& 0
& \cite{koklu2021classification}
& \cellcolor[rgb]{0.925, 0.957, 0.831}\textbf{94.93} (1)
& \cellcolor[rgb]{0.980,0.910,0.922}83.69 (2)
&  \cellcolor[rgb]{0.925, 0.957, 0.831}{\red{$^{\S}$}}\textbf{94.47} (1)
&\cellcolor[rgb]{0.925, 0.957, 0.831}\textbf{81.84} (1)
& \cellcolor[rgb]{0.925, 0.957, 0.831}\textbf{81.91} (1)
&  \cellcolor[rgb]{0.925, 0.957, 0.831}\textbf{83.18} (1)\\
Glass
& 214
& 9
& 6
& 0
& \cite{Dua:2019}
& 64.85 (2)
& \cellcolor[rgb]{0.980,0.910,0.922}43.90 (3)
&  \cellcolor[rgb]{0.925, 0.957, 0.831}{\red{$^{\dag\S}$}}\textbf{78.43} (1)
& 52.38 (2)
& \cellcolor[rgb]{0.980,0.910,0.922}37.62 (3)
&  \cellcolor[rgb]{0.925, 0.957, 0.831}{\red{$^{\dag\S}$}}\textbf{62.38} (1)\\
Hepatitis
& 155
& 19
& 2
& 5.67
& \cite{Dua:2019}
& \cellcolor[rgb]{0.925, 0.957, 0.831}\textbf{93.43} (1)
& \cellcolor[rgb]{0.980,0.910,0.922}89.67 (2)
&  \cellcolor[rgb]{0.925, 0.957, 0.831}{\red{$^{\S}$}}\textbf{92.55} (1)
& \cellcolor[rgb]{0.925, 0.957, 0.831}\textbf{50.89} (1)
& \cellcolor[rgb]{0.925, 0.957, 0.831}\textbf{49.56} (1)
& \cellcolor[rgb]{0.925, 0.957, 0.831}\textbf{52.00} (1)\\
Horse colic
& 368
& 22
& 2
& 23.80
& \cite{Dua:2019}
& \cellcolor[rgb]{0.925, 0.957, 0.831}\textbf{94.84} (1)
& \cellcolor[rgb]{0.980,0.910,0.922}85.03 (3)
&  {\blue{$^{\dag}$}}\hspace{-0.5pt}{\red{$^{\S}$}}90.37 (2)
& \cellcolor[rgb]{0.925, 0.957, 0.831}\textbf{61.39} (1)
& \cellcolor[rgb]{0.925, 0.957, 0.831}\textbf{62.13} (1)
& \cellcolor[rgb]{0.925, 0.957, 0.831}\textbf{59.63} (1)\\
Iris
& 150
& 4
& 3
& 0
& \cite{Dua:2019}
& \cellcolor[rgb]{0.980,0.910,0.922}86.89 (2)
& \cellcolor[rgb]{0.980,0.910,0.922}89.95 (2)
&  \cellcolor[rgb]{0.925, 0.957, 0.831}{\red{$^{\dag\S}$}}\textbf{95.60} (1)
& \cellcolor[rgb]{0.980,0.910,0.922}86.44 (2)
& \cellcolor[rgb]{0.980,0.910,0.922}90.22 (2)
&  \cellcolor[rgb]{0.925, 0.957, 0.831}{\red{$^{\dag\S}$}}\textbf{95.33} (1)\\
Mammographic masses
& 961
& 5
& 2
& 3.37
& \cite{Dua:2019}
& \cellcolor[rgb]{0.925, 0.957, 0.831}\textbf{62.91} (1)
& \cellcolor[rgb]{0.925, 0.957, 0.831}\textbf{60.17} (1)
& \cellcolor[rgb]{0.925, 0.957, 0.831}\textbf{60.91} (1)
& \cellcolor[rgb]{0.925, 0.957, 0.831}\textbf{58.68} (1)
& \cellcolor[rgb]{0.925, 0.957, 0.831}\textbf{62.15} (1)
& \cellcolor[rgb]{0.925, 0.957, 0.831}{\red{$^{\dag}$}}\textbf{62.43} (1)\\
Paddy leaf
& 6000
& 3
& 4
& 0
& \cite{wagner2022mechanisms}
& \cellcolor[rgb]{0.925, 0.957, 0.831}\textbf{91.50} (1)
& \cellcolor[rgb]{0.980,0.910,0.922}50.01 (2)
& \cellcolor[rgb]{0.925, 0.957, 0.831}{\red{$^{\S}$}}\textbf{91.77} (1)
& \cellcolor[rgb]{0.925, 0.957, 0.831}\textbf{89.71} (1)
& \cellcolor[rgb]{0.980,0.910,0.922}50.10 (2)
& \cellcolor[rgb]{0.925, 0.957, 0.831}{\red{$^{\dag\S}$}}\textbf{90.32} (1)\\
Pistachio
& 2148
& 16
& 2
& 0
& \cite{ozkan2021classification,singh2022classification}
& \cellcolor[rgb]{0.925, 0.957, 0.831}\textbf{87.83} (1)
& \cellcolor[rgb]{0.980,0.910,0.922}86.61 (3)
& {\red{$^{\S}$}}87.49 (2)
& \cellcolor[rgb]{0.925, 0.957, 0.831}\textbf{87.49} (1)
& \cellcolor[rgb]{0.925, 0.957, 0.831}\textbf{86.48} (1)
& \cellcolor[rgb]{0.925, 0.957, 0.831}\textbf{86.46} (1)\\
Raisin
& 900
& 7
& 2
& 0
&\cite{ccinar2020classification}
& 85.29 (2)
& \cellcolor[rgb]{0.980,0.910,0.922}83.29 (3)
&  \cellcolor[rgb]{0.925, 0.957, 0.831}{\red{$^{\dag\S}$}}\textbf{85.84} (1)
& \cellcolor[rgb]{0.925, 0.957, 0.831}\textbf{85.89} (1)
& \cellcolor[rgb]{0.980,0.910,0.922}82.33 (2)
&  \cellcolor[rgb]{0.925, 0.957, 0.831}{\red{$^{\S}$}}\textbf{84.81} (1)\\
Segment
& 2310
& 19
& 7
& 0
& \cite{Dua:2019}
& 96.36 (2)
& \cellcolor[rgb]{0.980,0.910,0.922}91.50 (3)
&\cellcolor[rgb]{0.925, 0.957, 0.831} {\red{$^{\dag\S}$}}\textbf{97.56} (1)
& 93.71 (2)
& \cellcolor[rgb]{0.980,0.910,0.922}90.30 (3)
&\cellcolor[rgb]{0.925, 0.957, 0.831} {\red{$^{\dag\S}$}}\textbf{95.56} (1)\\
Soybean
& 683
& 35
& 19
& 9.78
& \cite{Dua:2019}
& \cellcolor[rgb]{0.925, 0.957, 0.831}\textbf{98.24} (1)
& \cellcolor[rgb]{0.980,0.910,0.922}96.75 (3)
& {\blue{$^{\dag}$}}\hspace{-0.5pt}{\red{$^{\S}$}}97.47 (2)
& \cellcolor[rgb]{0.980,0.910,0.922}55.05 (3)
& \cellcolor[rgb]{0.925, 0.957, 0.831}\textbf{68.48} (1)
& {\red{$^{\dag}$}}\hspace{-0.5pt}{\blue{$^{\S}$}}59.90 (2)\\
Teaching assistant evaluation
& 151
& 5
& 3
& 0
& \cite{Dua:2019}
& 60.69 (2)
& \cellcolor[rgb]{0.980,0.910,0.922}53.58 (3)
& \cellcolor[rgb]{0.925, 0.957, 0.831}{\red{$^{\dag\S}$}}\textbf{67.40} (1)
& \cellcolor[rgb]{0.925, 0.957, 0.831}\textbf{53.33} (1)
& \cellcolor[rgb]{0.925, 0.957, 0.831}\textbf{47.11} (1)
& \cellcolor[rgb]{0.925, 0.957, 0.831}{\red{$^{\S}$}}\textbf{53.56} (1)\\
Wine
& 178
& 13
& 3
& 0
& \cite{Dua:2019}
& \cellcolor[rgb]{0.980,0.910,0.922}96.50 (2)
& \cellcolor[rgb]{0.980,0.910,0.922}96.89 (2)
& \cellcolor[rgb]{0.925, 0.957, 0.831}{\red{$^{\dag\S}$}}\textbf{98.96} (1)
& \cellcolor[rgb]{0.980,0.910,0.922}88.24 (2)
& \cellcolor[rgb]{0.925, 0.957, 0.831}\textbf{95.69} (1)
& \cellcolor[rgb]{0.925, 0.957, 0.831}{\red{$^{\S}$}}\textbf{96.27} (1)\\
Wisconsin breast-cancer
& 699
& 9
& 2
& 0.25
& \cite{Dua:2019}
& \cellcolor[rgb]{0.925, 0.957, 0.831}\textbf{97.06} (1)
& \cellcolor[rgb]{0.925, 0.957, 0.831}\textbf{96.97} (1)
& \cellcolor[rgb]{0.925, 0.957, 0.831}{\red{$^{\dag\S}$}}\textbf{97.10} (1)
& \cellcolor[rgb]{0.925, 0.957, 0.831}\textbf{94.25} (1)
& \cellcolor[rgb]{0.925, 0.957, 0.831}\textbf{95.31} (1)
& \cellcolor[rgb]{0.925, 0.957, 0.831}{\red{$^{\dag}$}}\textbf{95.51} (1)\\
Wisconsin prognostic breast-cancer
& 198
& 33
& 2
& 0.06
& \cite{Dua:2019}
& 97.58 (2)
& \cellcolor[rgb]{0.980,0.910,0.922}85.62 (3)
& \cellcolor[rgb]{0.925, 0.957, 0.831}{\red{$^{\dag\S}$}}\textbf{99.61} (1)
& \cellcolor[rgb]{0.925, 0.957, 0.831}\textbf{71.75} (1)
& \cellcolor[rgb]{0.925, 0.957, 0.831}\textbf{72.81} (1)
& \cellcolor[rgb]{0.925, 0.957, 0.831}{\blue{$^{\S}$}}\textbf{68.25} (1)\\
Yeast
& 1484
& 8
& 10
& 0
& \cite{Dua:2019}
& 58.51 (2)
& \cellcolor[rgb]{0.980,0.910,0.922}30.75 (3)
& \cellcolor[rgb]{0.925, 0.957, 0.831}{\red{$^{\dag\S}$}}\textbf{61.24} (1)
& \cellcolor[rgb]{0.925, 0.957, 0.831}\textbf{55.27} (1)
& \cellcolor[rgb]{0.980,0.910,0.922}28.33 (2)
& \cellcolor[rgb]{0.925, 0.957, 0.831}{\blue{$^{\dag}$}}\hspace{-0.5pt}{\red{$^{\S}$}}\textbf{53.63} (1)\\

\hline
\bhline{1pt}
\multicolumn{6}{r|}{\textit{Avg. Rank}}
&\textit{1.7}
&\textit{2.4}
&\multicolumn{1}{c|}{\textbf{\textit{1.2}}}
&\textit{1.5}
&\textit{1.5}
&\multicolumn{1}{c}{\textbf{\textit{1.1}}}\\
\hline
\multicolumn{6}{r|}{\multirow{2}{*}{\#\textit{Symbols}}}&
\multicolumn{2}{r}{\blue{$^{\dag}$}\black{UCS} - \red{$^{\dag}$}\black{Ours} : } & 2 - 13 & \multicolumn{2}{r}{\blue{$^{\dag}$}\black{UCS} - \red{$^{\dag}$}\black{Ours} : } & 1 - 11\\
&&&&&&\multicolumn{2}{r}{\blue{$^{\S}$}\black{Fuzzy-UCS} - \red{$^{\S}$}\black{Ours} : } & 0 - 19 & \multicolumn{2}{r}{\blue{$^{\S}$}\black{Fuzzy-UCS} - \red{$^{\S}$}\black{Ours} : } &2 -  12\\
\hline
\bhline{1pt}
\end{tabular}
}
\vspace{-4mm}
\end{center}
\end{table*}

\subsection{Discussion}
This section discusses experimental results for the RCB and NCB problems, where significant differences in the convergence classification accuracy among all systems were observed.
Figs. \ref{fig: rcb_all_landscape} and \ref{fig: ncb_all_landscape} depict the class \{0,1\} assignments inferred by the system for each coordinate input in the problem space, represented as black and white, respectively, following the completion of the 200,000 training steps for a certain seed (resolution 1000 $\times$ 1000).

In the RCB problem, as indicated in Table \ref{tb: result_exp1}, Adaptive-UCS demonstrates a significantly higher overall and convergence classification accuracy compared to UCS and Fuzzy-UCS. Upon closer examination of the area around the class boundary in Fig. \ref{fig: rcb_adaptiveucs_landscape}, it becomes apparent that Adaptive-UCS outperforms the other systems in approximating oblique class boundaries.
Fig. \ref{fig: rcb_ucs_landscape} illustrates that UCS approximates the oblique class boundaries through specific hyperrectangular rules, but its limitations in expressiveness result in unevenness observed on the boundaries. 
Fuzzy-UCS, depicted in Fig. \ref{fig: rcb_fuzzyucs_landscape}, utilizes hypertrapezoidal rules with $vague$ matching regions, thus providing a more accurate approximation of oblique class boundaries than UCS. 
However, the optimization difficulty caused by the more intricate rule structure in comparison to the other two systems is a hindrance in accurately approximating the majority of class boundary crossings. 
Our observation of this optimization difficulty is validated by the fact that the overall classification accuracy is significantly lower than UCS (as shown in Table \ref{tb: result_exp1}).
This can also be seen in Table. \ref{tb: result_exp1}, where the ruleset size is larger than the other two systems, resulting in redundant rule representations.
These findings are consistent with previous studies (e.g., \cite{lanzi2006using,wilson2008classifier}) that have investigated the impact of rule complexity on optimization performance.
In contrast, Fig. \ref{fig: rcb_adaptiveucs_landscape} illustrates that Adaptive-UCS provides a more accurate approximation of oblique class boundaries, as observed in the diamond-shaped checkerboard pattern in both the black and white regions.

The results of the NCB problem, as presented in Table \ref{tb: result_exp1} and Fig. \ref{fig: ncb_result}, highlight the limitations of using crisp rules in UCS. This is demonstrated by the lowest overall and convergence classification accuracy compared to other systems. This subpar performance is likely a result of overfitting of noisy inputs, as evident from the presence of white noise in the vicinity of class boundaries approximated by UCS, as depicted in Fig. \ref{fig: ncb_ucs_landscape}.
In essence, even if the rule representation for class boundaries is appropriate (e.g., the crisp-hyperrectangular representation is appropriate for class boundaries parallel to an axis), the presence of noise input significantly exacerbates the classification task using crisp rules.
Contrarily, the use of fuzzy rules in Fuzzy-UCS and Adaptive-UCS reveals the robustness of fuzzy systems to uncertainties such as noise \cite{casillas2007fuzzy,shoeleh2015knowledge}, as evidenced by the best convergence classification accuracy. The class boundaries approximated by these systems are noticeably devoid of white noise, further highlighting their robustness, as shown in Figs. \ref{fig: ncb_fuzzyucs_landscape} and \ref{fig: ncb_adaptiveucs_landscape}. However, it is noteworthy that the overall classification accuracy of Fuzzy-UCS is significantly lower than that of Adaptive-UCS, as shown in Table \ref{tb: result_exp1} and Fig. \ref{fig: ncb_result}. This can be attributed to the high optimization difficulty posed by the use of fuzzy-hypertrapezoidal rules, as discussed in the previous paragraph.

\section{Experiment 2: Real-World Problems}
\label{sec: experiment 2}

We employed 20 real-world datasets obtained from the \textit{UCI Repository} \cite{Dua:2019} and \textit{Kaggle Dataset}, as shown in Table \ref{tb:result}. The datasets possess inherent difficulties in data classification, such as missing values, class imbalance, and vague class boundaries \cite{nakata2018empirical}.
\subsection{Experimental Setup}

The experimental settings for all the systems remain unchanged from the CB and RCB problems in Experiment 1, except for the $r_0$ parameter, which governs the generality of the initial rule. 
In order to mitigate the risk of \textit{cover-delete} cycles \cite{butz2004toward}, a common occurrence in high-dimensional input problems \cite{tadokoro2021xcs,nakata2020learning}, the parameter is set to its maximum value of 1. 
The training phase lasts for 20 epochs, and each attribute value in the data is normalized to the range $[0, 1)$. The performance metric used to evaluate the systems is the average classification accuracy, obtained from 30 trials (3 iterations of 10-fold cross-validation) of both the training and test data. Additionally, the results are subjected to statistical analysis following the same procedure outlined in Experiment 1 (cf. Sect. \ref{ss: experiment1 setup}).

\subsection{Results and Discussion}
Table \ref{tb:result} presents each system's average classification accuracy and average rank across all training and test datasets.

The results of the experiments, as depicted in Table \ref{tb:result}, reveal that Adaptive-UCS outperforms both UCS and Fuzzy-UCS in terms of average rank. In particular, it outperforms UCS on 11 test datasets and Fuzzy-UCS on 12 test datasets, suggesting that Adaptive-UCS effectively averts the overfitting of rules, which can otherwise result in decreased classification accuracy during testing.
It is noteworthy that for some datasets with missing data attributes, such as Mammographic masses and Soybean, UCS demonstrates good performance during training but shows significant underperformance compared to Adaptive-UCS during testing. This could be attributed to the susceptibility of crisp rules to overfitting when dealing with highly uncertain inputs, such as missing inputs. Conversely, both Fuzzy-UCS and Adaptive-UCS, which incorporate fuzzy rules, exhibit weaker performance during training compared to UCS, but surpass it during testing. This aligns with the characteristics of fuzzy rules in being robust to uncertain and unknown inputs as discussed in previous works \cite{casillas2007fuzzy,shoeleh2015knowledge}.

However, Adaptive-UCS underperforms UCS on one dataset and Fuzzy-UCS on two datasets during testing, indicating that crisp rules were more effective in the former dataset and fuzzy rules were more effective in the latter datasets. Thus, there is still room for improvement in the self-adaptive rule representation mechanism of Adaptive-UCS. Currently, the mutation operator of the fuzzy indicator $\bm{\mathcal{F}}$  only alters the representation randomly. In contrast, a mutation incorporating stochastic local search has been proposed for XCS using ternary alphabet rule representation \cite{nakata2013simple}. It is suggested that introducing and extending the same mutation for Adaptive-UCS will enhance the classification performance by adding evolutionary pressure to perform a local search.

\section{Concluding Remarks}
\label{sec: concluding remarks}

In this paper, we focused on Fuzzy-UCS, proposed a self-adaptive rule representation mechanism, and named Fuzzy-UCS with this mechanism as Adaptive-UCS. 
Adaptive-UCS can attain optimal rule representation in the alphabet available to the system by optimizing the \textit{fuzzy indicator}, a critical parameter in the fuzzy rule representation, through evolutionary operations.
Our experiments on three benchmark problems  and 20 real-world data classification problems demonstrate that the self-adaptation of two simple rule membership functions (i.e., rectangular and triangular) can outperform the crisp-hyperrectangular and fuzzy-hypertrapezoidal rules. 
The results highlight the effectiveness of flexible rule representation by revisiting Fuzzy-UCS 15 years after its initial proposal in 2008. 
In conclusion, our findings provide evidence for the superiority of Adaptive-UCS in adapting rule representation and its effectiveness, especially in classifying uncertain data.

Although we applied the self-adaptation mechanism to UCS in this paper, the mechanism does not dictate the type of the correct answer label for the input data (e.g., nominal value indicating class, real value indicating function prediction, etc.) or its presence or absence. As such, it is also applicable to the XCSF classifier system \cite{wilson2002classifiers} for function prediction problems and XCS for reinforcement learning problems. Further investigation into the effectiveness of the mechanism in these LCSs is highly desirable.
Future works also include the extension of the recently proposed \textit{Evidential-UCS} framework \cite{ferjani2022evidential}, which employs the \textit{Dempster-Shafer theory} \cite{dempster1967upper} for addressing uncertainty in data classification, to Adaptive-UCS. These integration efforts hold the promise of enhancing classification performance to a further degree.

%
\bibliographystyle{ACM-Reference-Format}
\bibliography{camready}


\begin{thebibliography}{58}


\ifx \showCODEN    \undefined \def \showCODEN     #1{\unskip}     \fi
\ifx \showDOI      \undefined \def \showDOI       #1{#1}\fi
\ifx \showISBNx    \undefined \def \showISBNx     #1{\unskip}     \fi
\ifx \showISBNxiii \undefined \def \showISBNxiii  #1{\unskip}     \fi
\ifx \showISSN     \undefined \def \showISSN      #1{\unskip}     \fi
\ifx \showLCCN     \undefined \def \showLCCN      #1{\unskip}     \fi
\ifx \shownote     \undefined \def \shownote      #1{#1}          \fi
\ifx \showarticletitle \undefined \def \showarticletitle #1{#1}   \fi
\ifx \showURL      \undefined \def \showURL       {\relax}        \fi
\providecommand\bibfield[2]{#2}
\providecommand\bibinfo[2]{#2}
\providecommand\natexlab[1]{#1}
\providecommand\showeprint[2][]{arXiv:#2}

\bibitem[\protect\citeauthoryear{Arif, Li, and Iqbal}{Arif et~al\mbox{.}}{2017}]%
        {arif2017solving}
\bibfield{author}{\bibinfo{person}{Muhammad~Hassan Arif}, \bibinfo{person}{Jianxin Li}, {and} \bibinfo{person}{Muhammad Iqbal}.} \bibinfo{year}{2017}\natexlab{}.
\newblock \showarticletitle{Solving Social Media Text Classification Problems Using Code Fragment-Based XCSR}. In \bibinfo{booktitle}{\emph{2017 IEEE 29th International Conference on Tools with Artificial Intelligence (ICTAI)}}. \bibinfo{pages}{485--492}.
\newblock
\showISSN{2375-0197}
\urldef\tempurl%
\url{https://doi.org/10.1109/ICTAI.2017.00080}
\showDOI{\tempurl}


\bibitem[\protect\citeauthoryear{Bernad\'{o}-Mansilla and Garrell-Guiu}{Bernad\'{o}-Mansilla and Garrell-Guiu}{2003}]%
        {bernado2003accuracy}
\bibfield{author}{\bibinfo{person}{Ester Bernad\'{o}-Mansilla} {and} \bibinfo{person}{Josep~M. Garrell-Guiu}.} \bibinfo{year}{2003}\natexlab{}.
\newblock \showarticletitle{Accuracy-Based Learning Classifier Systems: Models, Analysis and Applications to Classification Tasks}.
\newblock \bibinfo{journal}{\emph{Evolutionary Computation}} \bibinfo{volume}{11}, \bibinfo{number}{3} (\bibinfo{date}{sep} \bibinfo{year}{2003}), \bibinfo{pages}{209–238}.
\newblock
\showISSN{1063-6560}
\urldef\tempurl%
\url{https://doi.org/10.1162/106365603322365289}
\showDOI{\tempurl}


\bibitem[\protect\citeauthoryear{Bishop and Gallagher}{Bishop and Gallagher}{2020}]%
        {bishop2020optimality}
\bibfield{author}{\bibinfo{person}{Jordan~T. Bishop} {and} \bibinfo{person}{Marcus Gallagher}.} \bibinfo{year}{2020}\natexlab{}.
\newblock \showarticletitle{Optimality-Based Analysis of {XCSF} Compaction in Discrete Reinforcement Learning}. In \bibinfo{booktitle}{\emph{Parallel Problem Solving from Nature – PPSN XVI: 16th International Conference, PPSN 2020, Leiden, The Netherlands, September 5-9, 2020, Proceedings, Part II}} (Leiden, The Netherlands). \bibinfo{publisher}{Springer-Verlag}, \bibinfo{address}{Berlin, Heidelberg}, \bibinfo{pages}{471–484}.
\newblock
\showISBNx{978-3-030-58114-5}
\urldef\tempurl%
\url{https://doi.org/10.1007/978-3-030-58115-2_33}
\showDOI{\tempurl}


\bibitem[\protect\citeauthoryear{Bishop, Gallagher, and Browne}{Bishop et~al\mbox{.}}{2021}]%
        {bishop2021genetic}
\bibfield{author}{\bibinfo{person}{Jordan~T. Bishop}, \bibinfo{person}{Marcus Gallagher}, {and} \bibinfo{person}{Will~N. Browne}.} \bibinfo{year}{2021}\natexlab{}.
\newblock \showarticletitle{A Genetic Fuzzy System for Interpretable and Parsimonious Reinforcement Learning Policies}. In \bibinfo{booktitle}{\emph{Proceedings of the Genetic and Evolutionary Computation Conference Companion}} (Lille, France) \emph{(\bibinfo{series}{GECCO '21})}. \bibinfo{publisher}{Association for Computing Machinery}, \bibinfo{address}{New York, NY, USA}, \bibinfo{pages}{1630–1638}.
\newblock
\showISBNx{9781450383516}
\urldef\tempurl%
\url{https://doi.org/10.1145/3449726.3463198}
\showDOI{\tempurl}


\bibitem[\protect\citeauthoryear{Bonarini}{Bonarini}{2000}]%
        {bonarini1999introduction}
\bibfield{author}{\bibinfo{person}{Andrea Bonarini}.} \bibinfo{year}{2000}\natexlab{}.
\newblock \showarticletitle{An Introduction to Learning Fuzzy Classifier Systems}. In \bibinfo{booktitle}{\emph{Learning Classifier Systems}}, \bibfield{editor}{\bibinfo{person}{Pier~Luca Lanzi}, \bibinfo{person}{Wolfgang Stolzmann}, {and} \bibinfo{person}{Stewart~W. Wilson}} (Eds.). \bibinfo{publisher}{Springer Berlin Heidelberg}, \bibinfo{address}{Berlin, Heidelberg}, \bibinfo{pages}{83--104}.
\newblock
\showISBNx{978-3-540-45027-6}
\urldef\tempurl%
\url{https://doi.org/10.1007/3-540-45027-0_4}
\showDOI{\tempurl}


\bibitem[\protect\citeauthoryear{Bull and O'Hara}{Bull and O'Hara}{2002}]%
        {bull2002accuracy}
\bibfield{author}{\bibinfo{person}{Larry Bull} {and} \bibinfo{person}{Toby O'Hara}.} \bibinfo{year}{2002}\natexlab{}.
\newblock \showarticletitle{Accuracy-Based Neuro and Neuro-Fuzzy Classifier Systems}. In \bibinfo{booktitle}{\emph{Proceedings of the 4th Annual Conference on Genetic and Evolutionary Computation}} (New York City, New York) \emph{(\bibinfo{series}{GECCO'02})}. \bibinfo{publisher}{Morgan Kaufmann Publishers Inc.}, \bibinfo{address}{San Francisco, CA, USA}, \bibinfo{pages}{905–911}.
\newblock
\showISBNx{1558608788}


\bibitem[\protect\citeauthoryear{Butz, Kovacs, Lanzi, and Wilson}{Butz et~al\mbox{.}}{2004}]%
        {butz2004toward}
\bibfield{author}{\bibinfo{person}{M.V. Butz}, \bibinfo{person}{T. Kovacs}, \bibinfo{person}{P.L. Lanzi}, {and} \bibinfo{person}{S.W. Wilson}.} \bibinfo{year}{2004}\natexlab{}.
\newblock \showarticletitle{Toward a theory of generalization and learning in {XCS}}.
\newblock \bibinfo{journal}{\emph{IEEE Transactions on Evolutionary Computation}} \bibinfo{volume}{8}, \bibinfo{number}{1} (\bibinfo{date}{Feb} \bibinfo{year}{2004}), \bibinfo{pages}{28--46}.
\newblock
\showISSN{1941-0026}
\urldef\tempurl%
\url{https://doi.org/10.1109/TEVC.2003.818194}
\showDOI{\tempurl}


\bibitem[\protect\citeauthoryear{Butz, Lanzi, and Wilson}{Butz et~al\mbox{.}}{2008}]%
        {butz2008function}
\bibfield{author}{\bibinfo{person}{Martin~V. Butz}, \bibinfo{person}{Pier~Luca Lanzi}, {and} \bibinfo{person}{Stewart~W. Wilson}.} \bibinfo{year}{2008}\natexlab{}.
\newblock \showarticletitle{Function Approximation With {XCS}: Hyperellipsoidal Conditions, Recursive Least Squares, and Compaction}.
\newblock \bibinfo{journal}{\emph{IEEE Transactions on Evolutionary Computation}} \bibinfo{volume}{12}, \bibinfo{number}{3} (\bibinfo{date}{June} \bibinfo{year}{2008}), \bibinfo{pages}{355--376}.
\newblock
\showISSN{1941-0026}
\urldef\tempurl%
\url{https://doi.org/10.1109/TEVC.2007.903551}
\showDOI{\tempurl}


\bibitem[\protect\citeauthoryear{Butz, Sastry, and Goldberg}{Butz et~al\mbox{.}}{2003}]%
        {butz2003tournament}
\bibfield{author}{\bibinfo{person}{Martin~V. Butz}, \bibinfo{person}{Kumara Sastry}, {and} \bibinfo{person}{David~E. Goldberg}.} \bibinfo{year}{2003}\natexlab{}.
\newblock \showarticletitle{Tournament Selection: Stable Fitness Pressure in {XCS}}. In \bibinfo{booktitle}{\emph{Genetic and Evolutionary Computation --- GECCO 2003}}. \bibinfo{publisher}{Springer Berlin Heidelberg}, \bibinfo{address}{Berlin, Heidelberg}, \bibinfo{pages}{1857--1869}.
\newblock
\showISBNx{978-3-540-45110-5}
\urldef\tempurl%
\url{https://doi.org/10.1007/3-540-45110-2_83}
\showDOI{\tempurl}


\bibitem[\protect\citeauthoryear{Casillas, Carse, and Bull}{Casillas et~al\mbox{.}}{2007}]%
        {casillas2007fuzzy}
\bibfield{author}{\bibinfo{person}{Jorge Casillas}, \bibinfo{person}{Brian Carse}, {and} \bibinfo{person}{Larry Bull}.} \bibinfo{year}{2007}\natexlab{}.
\newblock \showarticletitle{Fuzzy-{XCS}: A Michigan Genetic Fuzzy System}.
\newblock \bibinfo{journal}{\emph{IEEE Transactions on Fuzzy Systems}} \bibinfo{volume}{15}, \bibinfo{number}{4} (\bibinfo{date}{Aug} \bibinfo{year}{2007}), \bibinfo{pages}{536--550}.
\newblock
\showISSN{1941-0034}
\urldef\tempurl%
\url{https://doi.org/10.1109/TFUZZ.2007.900904}
\showDOI{\tempurl}


\bibitem[\protect\citeauthoryear{Chen, Douch, and Zhang}{Chen et~al\mbox{.}}{2016}]%
        {chen2016accuracy}
\bibfield{author}{\bibinfo{person}{Gang Chen}, \bibinfo{person}{Colin I.~J. Douch}, {and} \bibinfo{person}{Mengjie Zhang}.} \bibinfo{year}{2016}\natexlab{}.
\newblock \showarticletitle{Accuracy-Based Learning Classifier Systems for Multistep Reinforcement Learning: A Fuzzy Logic Approach to Handling Continuous Inputs and Learning Continuous Actions}.
\newblock \bibinfo{journal}{\emph{IEEE Transactions on Evolutionary Computation}} \bibinfo{volume}{20}, \bibinfo{number}{6} (\bibinfo{date}{Dec} \bibinfo{year}{2016}), \bibinfo{pages}{953--971}.
\newblock
\showISSN{1941-0026}
\urldef\tempurl%
\url{https://doi.org/10.1109/TEVC.2016.2560139}
\showDOI{\tempurl}


\bibitem[\protect\citeauthoryear{{\c{C}}inar, Koklu, and Ta{\c{s}}dem{\.i}r}{{\c{C}}inar et~al\mbox{.}}{2020}]%
        {ccinar2020classification}
\bibfield{author}{\bibinfo{person}{{\.I}lkay {\c{C}}inar}, \bibinfo{person}{Murat Koklu}, {and} \bibinfo{person}{{\c{S}}akir Ta{\c{s}}dem{\.i}r}.} \bibinfo{year}{2020}\natexlab{}.
\newblock \showarticletitle{Classification of raisin grains using machine vision and artificial intelligence methods}.
\newblock \bibinfo{journal}{\emph{Gazi M{\"u}hendislik Bilimleri Dergisi}} \bibinfo{volume}{6}, \bibinfo{number}{3} (\bibinfo{date}{12} \bibinfo{year}{2020}), \bibinfo{pages}{200--209}.
\newblock
\urldef\tempurl%
\url{https://doi.org/10.30855/gmbd.2020.03.03}
\showDOI{\tempurl}


\bibitem[\protect\citeauthoryear{Dam, Abbass, and Lokan}{Dam et~al\mbox{.}}{2005}]%
        {dam2005real}
\bibfield{author}{\bibinfo{person}{Hai~H. Dam}, \bibinfo{person}{Hussein~A. Abbass}, {and} \bibinfo{person}{Chris Lokan}.} \bibinfo{year}{2005}\natexlab{}.
\newblock \showarticletitle{Be real! {XCS} with continuous-valued inputs}. In \bibinfo{booktitle}{\emph{Proceedings of the 7th Annual Workshop on Genetic and Evolutionary Computation}} (Washington, D.C.) \emph{(\bibinfo{series}{GECCO '05})}. \bibinfo{publisher}{Association for Computing Machinery}, \bibinfo{address}{New York, NY, USA}, \bibinfo{pages}{85–87}.
\newblock
\showISBNx{9781450378000}
\urldef\tempurl%
\url{https://doi.org/10.1145/1102256.1102274}
\showDOI{\tempurl}


\bibitem[\protect\citeauthoryear{Dempster}{Dempster}{1967}]%
        {dempster1967upper}
\bibfield{author}{\bibinfo{person}{Arthur~P. Dempster}.} \bibinfo{year}{1967}\natexlab{}.
\newblock \showarticletitle{{Upper and Lower Probabilities Induced by a Multivalued Mapping}}.
\newblock \bibinfo{journal}{\emph{The Annals of Mathematical Statistics}} \bibinfo{volume}{38}, \bibinfo{number}{2} (\bibinfo{year}{1967}), \bibinfo{pages}{325 -- 339}.
\newblock
\urldef\tempurl%
\url{https://doi.org/10.1214/aoms/1177698950}
\showDOI{\tempurl}


\bibitem[\protect\citeauthoryear{Dua and Graff}{Dua and Graff}{2017}]%
        {Dua:2019}
\bibfield{author}{\bibinfo{person}{Dheeru Dua} {and} \bibinfo{person}{Casey Graff}.} \bibinfo{year}{2017}\natexlab{}.
\newblock \bibinfo{title}{{UCI} Machine Learning Repository}.
\newblock
\newblock
\urldef\tempurl%
\url{http://archive.ics.uci.edu/ml}
\showURL{%
\tempurl}


\bibitem[\protect\citeauthoryear{Ferjani, Rejeb, Abdelkarim, and Said}{Ferjani et~al\mbox{.}}{2022}]%
        {ferjani2022evidential}
\bibfield{author}{\bibinfo{person}{Rahma Ferjani}, \bibinfo{person}{Lilia Rejeb}, \bibinfo{person}{Chedi Abdelkarim}, {and} \bibinfo{person}{Lamjed~Ben Said}.} \bibinfo{year}{2022}\natexlab{}.
\newblock \showarticletitle{Evidential Supervised Classifier System: A New Learning Classifier System Dealing with Imperfect Information}.
\newblock \bibinfo{journal}{\emph{International Journal of Information Technology \& Decision Making}} (\bibinfo{year}{2022}), \bibinfo{pages}{1--22}.
\newblock
\urldef\tempurl%
\url{https://doi.org/10.1142/S0219622022500997}
\showDOI{\tempurl}


\bibitem[\protect\citeauthoryear{Goldberg}{Goldberg}{1989}]%
        {goldberg1989genetic}
\bibfield{author}{\bibinfo{person}{David~E. Goldberg}.} \bibinfo{year}{1989}\natexlab{}.
\newblock \bibinfo{booktitle}{\emph{Genetic Algorithms in Search, Optimization and Machine Learning} (\bibinfo{edition}{1st} ed.)}.
\newblock \bibinfo{publisher}{Addison-Wesley Longman Publishing Co., Inc.}, \bibinfo{address}{USA}.
\newblock
\showISBNx{0201157675}


\bibitem[\protect\citeauthoryear{Guevara and Santos}{Guevara and Santos}{2021}]%
        {guevara2021intelligent}
\bibfield{author}{\bibinfo{person}{C{\'e}sar Guevara} {and} \bibinfo{person}{Matilde Santos}.} \bibinfo{year}{2021}\natexlab{}.
\newblock \showarticletitle{Intelligent models for movement detection and physical evolution of patients with hip surgery}.
\newblock \bibinfo{journal}{\emph{Logic Journal of the IGPL}} \bibinfo{volume}{29}, \bibinfo{number}{6} (\bibinfo{year}{2021}), \bibinfo{pages}{874--888}.
\newblock
\showISSN{1367-0751}
\urldef\tempurl%
\url{https://doi.org/10.1093/jigpal/jzaa032}
\showDOI{\tempurl}


\bibitem[\protect\citeauthoryear{Heider, Stegherr, Wurth, Sraj, and H{\"a}hner}{Heider et~al\mbox{.}}{2022}]%
        {heider2022investigating}
\bibfield{author}{\bibinfo{person}{Michael Heider}, \bibinfo{person}{Helena Stegherr}, \bibinfo{person}{Jonathan Wurth}, \bibinfo{person}{Roman Sraj}, {and} \bibinfo{person}{J{\"o}rg H{\"a}hner}.} \bibinfo{year}{2022}\natexlab{}.
\newblock \showarticletitle{Investigating the Impact of Independent Rule Fitnesses in a Learning Classifier System}. In \bibinfo{booktitle}{\emph{Bioinspired Optimization Methods and Their Applications}}. \bibinfo{publisher}{Springer International Publishing}, \bibinfo{address}{Cham}, \bibinfo{pages}{142--156}.
\newblock
\showISBNx{978-3-031-21094-5}
\urldef\tempurl%
\url{https://doi.org/10.1007/978-3-031-21094-5_11}
\showDOI{\tempurl}


\bibitem[\protect\citeauthoryear{Holland}{Holland}{1986}]%
        {holland1986possibilities}
\bibfield{author}{\bibinfo{person}{John~H Holland}.} \bibinfo{year}{1986}\natexlab{}.
\newblock \showarticletitle{Escaping Brittleness: The possibilities of general-purpose learning algorithms applied to parallel rule-based systems}.
\newblock \bibinfo{journal}{\emph{Machine learning, an artificial intelligence approach}}  \bibinfo{volume}{2} (\bibinfo{year}{1986}), \bibinfo{pages}{593--623}.
\newblock


\bibitem[\protect\citeauthoryear{Iqbal, Browne, and Zhang}{Iqbal et~al\mbox{.}}{2014}]%
        {iqbal2013reusing}
\bibfield{author}{\bibinfo{person}{Muhammad Iqbal}, \bibinfo{person}{Will~N. Browne}, {and} \bibinfo{person}{Mengjie Zhang}.} \bibinfo{year}{2014}\natexlab{}.
\newblock \showarticletitle{Reusing Building Blocks of Extracted Knowledge to Solve Complex, Large-Scale Boolean Problems}.
\newblock \bibinfo{journal}{\emph{IEEE Transactions on Evolutionary Computation}} \bibinfo{volume}{18}, \bibinfo{number}{4} (\bibinfo{year}{2014}), \bibinfo{pages}{465--480}.
\newblock
\urldef\tempurl%
\url{https://doi.org/10.1109/TEVC.2013.2281537}
\showDOI{\tempurl}


\bibitem[\protect\citeauthoryear{Ishibuchi and Yamamoto}{Ishibuchi and Yamamoto}{2005}]%
        {ishibuchi2005rule}
\bibfield{author}{\bibinfo{person}{Hisao Ishibuchi} {and} \bibinfo{person}{Takashi Yamamoto}.} \bibinfo{year}{2005}\natexlab{}.
\newblock \showarticletitle{Rule weight specification in fuzzy rule-based classification systems}.
\newblock \bibinfo{journal}{\emph{IEEE Transactions on Fuzzy Systems}} \bibinfo{volume}{13}, \bibinfo{number}{4} (\bibinfo{date}{Aug} \bibinfo{year}{2005}), \bibinfo{pages}{428--435}.
\newblock
\showISSN{1941-0034}
\urldef\tempurl%
\url{https://doi.org/10.1109/TFUZZ.2004.841738}
\showDOI{\tempurl}


\bibitem[\protect\citeauthoryear{Koklu, Kursun, Taspinar, and Cinar}{Koklu et~al\mbox{.}}{2021}]%
        {koklu2021classification}
\bibfield{author}{\bibinfo{person}{Murat Koklu}, \bibinfo{person}{Ramazan Kursun}, \bibinfo{person}{Yavuz~Selim Taspinar}, {and} \bibinfo{person}{Ilkay Cinar}.} \bibinfo{year}{2021}\natexlab{}.
\newblock \showarticletitle{Classification of date fruits into genetic varieties using image analysis}.
\newblock \bibinfo{journal}{\emph{Mathematical Problems in Engineering}}  \bibinfo{volume}{2021} (\bibinfo{year}{2021}).
\newblock
\urldef\tempurl%
\url{https://doi.org/10.1155/2021/4793293}
\showDOI{\tempurl}


\bibitem[\protect\citeauthoryear{Lanzi and Wilson}{Lanzi and Wilson}{2006}]%
        {lanzi2006using}
\bibfield{author}{\bibinfo{person}{Pier~Luca Lanzi} {and} \bibinfo{person}{Stewart~W. Wilson}.} \bibinfo{year}{2006}\natexlab{}.
\newblock \showarticletitle{Using Convex Hulls to Represent Classifier Conditions}. In \bibinfo{booktitle}{\emph{Proceedings of the 8th Annual Conference on Genetic and Evolutionary Computation}} (Seattle, Washington, USA) \emph{(\bibinfo{series}{GECCO '06})}. \bibinfo{publisher}{Association for Computing Machinery}, \bibinfo{address}{New York, NY, USA}, \bibinfo{pages}{1481–1488}.
\newblock
\showISBNx{1595931864}
\urldef\tempurl%
\url{https://doi.org/10.1145/1143997.1144240}
\showDOI{\tempurl}


\bibitem[\protect\citeauthoryear{Liu, Browne, and Xue}{Liu et~al\mbox{.}}{2020}]%
        {liu2020absumption}
\bibfield{author}{\bibinfo{person}{Yi Liu}, \bibinfo{person}{Will~N. Browne}, {and} \bibinfo{person}{Bing Xue}.} \bibinfo{year}{2020}\natexlab{}.
\newblock \showarticletitle{Absumption and Subsumption Based Learning Classifier Systems}. In \bibinfo{booktitle}{\emph{Proceedings of the 2020 Genetic and Evolutionary Computation Conference}} (Canc\'{u}n, Mexico) \emph{(\bibinfo{series}{GECCO '20})}. \bibinfo{publisher}{Association for Computing Machinery}, \bibinfo{address}{New York, NY, USA}, \bibinfo{pages}{368–376}.
\newblock
\showISBNx{9781450371285}
\urldef\tempurl%
\url{https://doi.org/10.1145/3377930.3389813}
\showDOI{\tempurl}


\bibitem[\protect\citeauthoryear{Nakata and Browne}{Nakata and Browne}{2021}]%
        {nakata2020learning}
\bibfield{author}{\bibinfo{person}{Masaya Nakata} {and} \bibinfo{person}{Will~N. Browne}.} \bibinfo{year}{2021}\natexlab{}.
\newblock \showarticletitle{Learning Optimality Theory for Accuracy-Based Learning Classifier Systems}.
\newblock \bibinfo{journal}{\emph{IEEE Transactions on Evolutionary Computation}} \bibinfo{volume}{25}, \bibinfo{number}{1} (\bibinfo{date}{Feb} \bibinfo{year}{2021}), \bibinfo{pages}{61--74}.
\newblock
\showISSN{1941-0026}
\urldef\tempurl%
\url{https://doi.org/10.1109/TEVC.2020.2994314}
\showDOI{\tempurl}


\bibitem[\protect\citeauthoryear{Nakata, Lanzi, and Takadama}{Nakata et~al\mbox{.}}{2013}]%
        {nakata2013simple}
\bibfield{author}{\bibinfo{person}{Masaya Nakata}, \bibinfo{person}{Pier~Luca Lanzi}, {and} \bibinfo{person}{Keiki Takadama}.} \bibinfo{year}{2013}\natexlab{}.
\newblock \showarticletitle{Simple compact genetic algorithm for XCS}. In \bibinfo{booktitle}{\emph{2013 IEEE Congress on Evolutionary Computation}}. \bibinfo{pages}{1718--1723}.
\newblock
\showISSN{1941-0026}
\urldef\tempurl%
\url{https://doi.org/10.1109/CEC.2013.6557768}
\showDOI{\tempurl}


\bibitem[\protect\citeauthoryear{Nakata and Takadama}{Nakata and Takadama}{2018}]%
        {nakata2018empirical}
\bibfield{author}{\bibinfo{person}{Masaya Nakata} {and} \bibinfo{person}{Keiki Takadama}.} \bibinfo{year}{2018}\natexlab{}.
\newblock \showarticletitle{An Empirical Analysis of Action Map in Learning Classifier Systems}.
\newblock \bibinfo{journal}{\emph{SICE Journal of Control, Measurement, and System Integration}} \bibinfo{volume}{11}, \bibinfo{number}{3} (\bibinfo{year}{2018}), \bibinfo{pages}{239--248}.
\newblock
\urldef\tempurl%
\url{https://doi.org/10.9746/jcmsi.11.239}
\showDOI{\tempurl}


\bibitem[\protect\citeauthoryear{Orriols-Puig and Casillas}{Orriols-Puig and Casillas}{2011}]%
        {orriols2011fuzzy}
\bibfield{author}{\bibinfo{person}{Albert Orriols-Puig} {and} \bibinfo{person}{Jorge Casillas}.} \bibinfo{year}{2011}\natexlab{}.
\newblock \showarticletitle{Fuzzy knowledge representation study for incremental learning in data streams and classification problems}.
\newblock \bibinfo{journal}{\emph{Soft Computing}} \bibinfo{volume}{15}, \bibinfo{number}{12} (\bibinfo{year}{2011}), \bibinfo{pages}{2389--2414}.
\newblock
\urldef\tempurl%
\url{https://doi.org/10.1007/s00500-010-0668-x}
\showDOI{\tempurl}


\bibitem[\protect\citeauthoryear{Orriols-Puig, Casillas, and Bernad{\'o}-Mansilla}{Orriols-Puig et~al\mbox{.}}{2008a}]%
        {orriols2008approximate}
\bibfield{author}{\bibinfo{person}{Albert Orriols-Puig}, \bibinfo{person}{Jorge Casillas}, {and} \bibinfo{person}{Ester Bernad{\'o}-Mansilla}.} \bibinfo{year}{2008}\natexlab{a}.
\newblock \showarticletitle{Approximate Versus Linguistic Representation in Fuzzy-{UCS}}. In \bibinfo{booktitle}{\emph{Hybrid Artificial Intelligence Systems}}. \bibinfo{publisher}{Springer Berlin Heidelberg}, \bibinfo{address}{Berlin, Heidelberg}, \bibinfo{pages}{722--729}.
\newblock
\showISBNx{978-3-540-87656-4}
\urldef\tempurl%
\url{https://doi.org/10.1007/978-3-540-87656-4_89}
\showDOI{\tempurl}


\bibitem[\protect\citeauthoryear{Orriols-Puig, Casillas, and Bernad{\'o}-Mansilla}{Orriols-Puig et~al\mbox{.}}{2008b}]%
        {orriols2008evolving}
\bibfield{author}{\bibinfo{person}{Albert Orriols-Puig}, \bibinfo{person}{Jorge Casillas}, {and} \bibinfo{person}{Ester Bernad{\'o}-Mansilla}.} \bibinfo{year}{2008}\natexlab{b}.
\newblock \showarticletitle{Evolving Fuzzy Rules with {UCS}: Preliminary Results}. In \bibinfo{booktitle}{\emph{Learning Classifier Systems}}. \bibinfo{publisher}{Springer Berlin Heidelberg}, \bibinfo{address}{Berlin, Heidelberg}, \bibinfo{pages}{57--76}.
\newblock
\showISBNx{978-3-540-88138-4}
\urldef\tempurl%
\url{https://doi.org/10.1007/978-3-540-88138-4_4}
\showDOI{\tempurl}


\bibitem[\protect\citeauthoryear{Orriols-Puig, Casillas, and Bernad{\'o}-Mansilla}{Orriols-Puig et~al\mbox{.}}{2009}]%
        {orriols2008fuzzy}
\bibfield{author}{\bibinfo{person}{Albert Orriols-Puig}, \bibinfo{person}{Jorge Casillas}, {and} \bibinfo{person}{Ester Bernad{\'o}-Mansilla}.} \bibinfo{year}{2009}\natexlab{}.
\newblock \showarticletitle{Fuzzy-{UCS}: A Michigan-Style Learning Fuzzy-Classifier System for Supervised Learning}.
\newblock \bibinfo{journal}{\emph{IEEE Transactions on Evolutionary Computation}} \bibinfo{volume}{13}, \bibinfo{number}{2} (\bibinfo{date}{April} \bibinfo{year}{2009}), \bibinfo{pages}{260--283}.
\newblock
\showISSN{1941-0026}
\urldef\tempurl%
\url{https://doi.org/10.1109/TEVC.2008.925144}
\showDOI{\tempurl}


\bibitem[\protect\citeauthoryear{{\"O}zkan, K{\"o}kl{\"u}, and Sara{\c{c}}o{\u{g}}lu}{{\"O}zkan et~al\mbox{.}}{2021}]%
        {ozkan2021classification}
\bibfield{author}{\bibinfo{person}{{\.I}lker {\"O}zkan}, \bibinfo{person}{Murat K{\"o}kl{\"u}}, {and} \bibinfo{person}{R{\i}dvan Sara{\c{c}}o{\u{g}}lu}.} \bibinfo{year}{2021}\natexlab{}.
\newblock \showarticletitle{Classification of Pistachio Species Using Improved k-NN Classifier}.
\newblock \bibinfo{journal}{\emph{Progress in Nutrition}} \bibinfo{volume}{23}, \bibinfo{number}{2} (\bibinfo{year}{2021}).
\newblock
\urldef\tempurl%
\url{https://doi.org/10.23751/pn.v23i2.9686}
\showDOI{\tempurl}


\bibitem[\protect\citeauthoryear{Shiraishi, Hayamizu, Sato, and Takadama}{Shiraishi et~al\mbox{.}}{2022a}]%
        {shiraishi2022absumption}
\bibfield{author}{\bibinfo{person}{Hiroki Shiraishi}, \bibinfo{person}{Yohei Hayamizu}, \bibinfo{person}{Hiroyuki Sato}, {and} \bibinfo{person}{Keiki Takadama}.} \bibinfo{year}{2022}\natexlab{a}.
\newblock \showarticletitle{Absumption Based on Overgenerality and Condition-Clustering Based Specialization for {XCS} with Continuous-Valued Inputs}. In \bibinfo{booktitle}{\emph{Proceedings of the Genetic and Evolutionary Computation Conference}} (Boston, Massachusetts) \emph{(\bibinfo{series}{GECCO '22})}. \bibinfo{publisher}{Association for Computing Machinery}, \bibinfo{address}{New York, NY, USA}, \bibinfo{pages}{422–430}.
\newblock
\showISBNx{9781450392372}
\urldef\tempurl%
\url{https://doi.org/10.1145/3512290.3528841}
\showDOI{\tempurl}


\bibitem[\protect\citeauthoryear{Shiraishi, Hayamizu, Sato, and Takadama}{Shiraishi et~al\mbox{.}}{2022b}]%
        {shiraishi2022beta}
\bibfield{author}{\bibinfo{person}{Hiroki Shiraishi}, \bibinfo{person}{Yohei Hayamizu}, \bibinfo{person}{Hiroyuki Sato}, {and} \bibinfo{person}{Keiki Takadama}.} \bibinfo{year}{2022}\natexlab{b}.
\newblock \showarticletitle{Beta Distribution based {XCS} Classifier System}. In \bibinfo{booktitle}{\emph{2022 IEEE Congress on Evolutionary Computation (CEC)}}. \bibinfo{publisher}{IEEE}, \bibinfo{pages}{1--8}.
\newblock
\urldef\tempurl%
\url{https://doi.org/10.1109/CEC55065.2022.9870314}
\showDOI{\tempurl}


\bibitem[\protect\citeauthoryear{Shiraishi, Hayamizu, Sato, and Takadama}{Shiraishi et~al\mbox{.}}{2022c}]%
        {shiraishi2022can}
\bibfield{author}{\bibinfo{person}{Hiroki Shiraishi}, \bibinfo{person}{Yohei Hayamizu}, \bibinfo{person}{Hiroyuki Sato}, {and} \bibinfo{person}{Keiki Takadama}.} \bibinfo{year}{2022}\natexlab{c}.
\newblock \showarticletitle{Can the Same Rule Representation Change Its Matching Area? Enhancing Representation in {XCS} for Continuous Space by Probability Distribution in Multiple Dimension}. In \bibinfo{booktitle}{\emph{Proceedings of the Genetic and Evolutionary Computation Conference}} (Boston, Massachusetts) \emph{(\bibinfo{series}{GECCO '22})}. \bibinfo{publisher}{Association for Computing Machinery}, \bibinfo{address}{New York, NY, USA}, \bibinfo{pages}{431–439}.
\newblock
\showISBNx{9781450392372}
\urldef\tempurl%
\url{https://doi.org/10.1145/3512290.3528874}
\showDOI{\tempurl}


\bibitem[\protect\citeauthoryear{Shoeleh, Hamzeh, and Hashemi}{Shoeleh et~al\mbox{.}}{2010}]%
        {shoeleh2010handle}
\bibfield{author}{\bibinfo{person}{Farzaneh Shoeleh}, \bibinfo{person}{Ali Hamzeh}, {and} \bibinfo{person}{Sattar Hashemi}.} \bibinfo{year}{2010}\natexlab{}.
\newblock \showarticletitle{To Handle Real Valued Input in {XCS}: Using Fuzzy Hyper-trapezoidal Membership in Classifier Condition}. In \bibinfo{booktitle}{\emph{Simulated Evolution and Learning}}. \bibinfo{publisher}{Springer Berlin Heidelberg}, \bibinfo{address}{Berlin, Heidelberg}, \bibinfo{pages}{55--64}.
\newblock
\showISBNx{978-3-642-17298-4}
\urldef\tempurl%
\url{https://doi.org/10.1007/978-3-642-17298-4_5}
\showDOI{\tempurl}


\bibitem[\protect\citeauthoryear{Shoeleh, Hamzeh, and Hashemi}{Shoeleh et~al\mbox{.}}{2011}]%
        {shoeleh2011towards}
\bibfield{author}{\bibinfo{person}{Farzaneh Shoeleh}, \bibinfo{person}{Ali Hamzeh}, {and} \bibinfo{person}{Sattar Hashemi}.} \bibinfo{year}{2011}\natexlab{}.
\newblock \showarticletitle{Towards Final Rule Set Reduction in {XCS}: A Fuzzy Representation Approach}. In \bibinfo{booktitle}{\emph{Proceedings of the 13th Annual Conference on Genetic and Evolutionary Computation}} (Dublin, Ireland) \emph{(\bibinfo{series}{GECCO '11})}. \bibinfo{publisher}{Association for Computing Machinery}, \bibinfo{address}{New York, NY, USA}, \bibinfo{pages}{1211–1218}.
\newblock
\showISBNx{9781450305570}
\urldef\tempurl%
\url{https://doi.org/10.1145/2001576.2001740}
\showDOI{\tempurl}


\bibitem[\protect\citeauthoryear{Shoeleh, Majd, Hamzeh, and Hashemi}{Shoeleh et~al\mbox{.}}{2015}]%
        {shoeleh2015knowledge}
\bibfield{author}{\bibinfo{person}{Farzaneh Shoeleh}, \bibinfo{person}{Mahshid Majd}, \bibinfo{person}{Ali Hamzeh}, {and} \bibinfo{person}{Sattar Hashemi}.} \bibinfo{year}{2015}\natexlab{}.
\newblock \bibinfo{title}{Knowledge Representation in Learning Classifier Systems: A Review}.
\newblock
\newblock
\urldef\tempurl%
\url{https://doi.org/10.48550/ARXIV.1506.04002}
\showDOI{\tempurl}


\bibitem[\protect\citeauthoryear{Singh, Taspinar, Kursun, Cinar, Koklu, Ozkan, and Lee}{Singh et~al\mbox{.}}{2022}]%
        {singh2022classification}
\bibfield{author}{\bibinfo{person}{Dilbag Singh}, \bibinfo{person}{Yavuz~Selim Taspinar}, \bibinfo{person}{Ramazan Kursun}, \bibinfo{person}{Ilkay Cinar}, \bibinfo{person}{Murat Koklu}, \bibinfo{person}{Ilker~Ali Ozkan}, {and} \bibinfo{person}{Heung-No Lee}.} \bibinfo{year}{2022}\natexlab{}.
\newblock \showarticletitle{Classification and Analysis of Pistachio Species with Pre-Trained Deep Learning Models}.
\newblock \bibinfo{journal}{\emph{Electronics}} \bibinfo{volume}{11}, \bibinfo{number}{7} (\bibinfo{year}{2022}).
\newblock
\showISSN{2079-9292}
\urldef\tempurl%
\url{https://doi.org/10.3390/electronics11070981}
\showDOI{\tempurl}


\bibitem[\protect\citeauthoryear{Smith, Everhart, Dickson, Knowler, and Johannes}{Smith et~al\mbox{.}}{1988}]%
        {smith1988using}
\bibfield{author}{\bibinfo{person}{Jack~W Smith}, \bibinfo{person}{James~E Everhart}, \bibinfo{person}{WC Dickson}, \bibinfo{person}{William~C Knowler}, {and} \bibinfo{person}{Robert~Scott Johannes}.} \bibinfo{year}{1988}\natexlab{}.
\newblock \showarticletitle{Using the {ADAP} learning algorithm to forecast the onset of diabetes mellitus}. In \bibinfo{booktitle}{\emph{Proceedings of the annual symposium on computer application in medical care}}. American Medical Informatics Association, \bibinfo{pages}{261}.
\newblock
\urldef\tempurl%
\url{https://europepmc.org/articles/PMC2245318}
\showURL{%
\tempurl}


\bibitem[\protect\citeauthoryear{Stone and Bull}{Stone and Bull}{2003}]%
        {stone2003real}
\bibfield{author}{\bibinfo{person}{Christopher Stone} {and} \bibinfo{person}{Larry Bull}.} \bibinfo{year}{2003}\natexlab{}.
\newblock \showarticletitle{For Real! {XCS} with Continuous-Valued Inputs}.
\newblock \bibinfo{journal}{\emph{Evol. Comput.}} \bibinfo{volume}{11}, \bibinfo{number}{3} (\bibinfo{date}{sep} \bibinfo{year}{2003}), \bibinfo{pages}{299–336}.
\newblock
\showISSN{1063-6560}
\urldef\tempurl%
\url{https://doi.org/10.1162/106365603322365315}
\showDOI{\tempurl}


\bibitem[\protect\citeauthoryear{Tadokoro, Sato, and Takadama}{Tadokoro et~al\mbox{.}}{2021}]%
        {tadokoro2021xcs}
\bibfield{author}{\bibinfo{person}{Masakazu Tadokoro}, \bibinfo{person}{Hiroyuki Sato}, {and} \bibinfo{person}{Keiki Takadama}.} \bibinfo{year}{2021}\natexlab{}.
\newblock \showarticletitle{{XCS} with Weight-based Matching in VAE Latent Space and Additional Learning of High-Dimensional Data}. In \bibinfo{booktitle}{\emph{2021 IEEE Congress on Evolutionary Computation (CEC)}}. \bibinfo{publisher}{IEEE}, \bibinfo{pages}{304--310}.
\newblock
\urldef\tempurl%
\url{https://doi.org/10.1109/CEC45853.2021.9504909}
\showDOI{\tempurl}


\bibitem[\protect\citeauthoryear{Urbanowicz, Andrew, Karagas, and Moore}{Urbanowicz et~al\mbox{.}}{2013}]%
        {urbanowicz2013role}
\bibfield{author}{\bibinfo{person}{Ryan~John Urbanowicz}, \bibinfo{person}{Angeline~S Andrew}, \bibinfo{person}{Margaret~Rita Karagas}, {and} \bibinfo{person}{Jason~H Moore}.} \bibinfo{year}{2013}\natexlab{}.
\newblock \showarticletitle{{Role of genetic heterogeneity and epistasis in bladder cancer susceptibility and outcome: a learning classifier system approach}}.
\newblock \bibinfo{journal}{\emph{Journal of the American Medical Informatics Association}} \bibinfo{volume}{20}, \bibinfo{number}{4} (\bibinfo{date}{02} \bibinfo{year}{2013}), \bibinfo{pages}{603--612}.
\newblock
\showISSN{1067-5027}
\urldef\tempurl%
\url{https://doi.org/10.1136/amiajnl-2012-001574}
\showDOI{\tempurl}


\bibitem[\protect\citeauthoryear{Urbanowicz and Browne}{Urbanowicz and Browne}{2017}]%
        {urbanowicz2017introduction}
\bibfield{author}{\bibinfo{person}{Ryan~John Urbanowicz} {and} \bibinfo{person}{Will~N. Browne}.} \bibinfo{year}{2017}\natexlab{}.
\newblock \bibinfo{booktitle}{\emph{Introduction to Learning Classifier Systems} (\bibinfo{edition}{1st} ed.)}.
\newblock \bibinfo{publisher}{Springer Publishing Company, Incorporated}.
\newblock
\showISBNx{3662550067}


\bibitem[\protect\citeauthoryear{Urbanowicz and Moore}{Urbanowicz and Moore}{2009}]%
        {urbanowicz2009learning}
\bibfield{author}{\bibinfo{person}{Ryan~John Urbanowicz} {and} \bibinfo{person}{Jason~H Moore}.} \bibinfo{year}{2009}\natexlab{}.
\newblock \showarticletitle{Learning classifier systems: a complete introduction, review, and roadmap}.
\newblock \bibinfo{journal}{\emph{Journal of Artificial Evolution and Applications}}  \bibinfo{volume}{2009} (\bibinfo{year}{2009}).
\newblock
\urldef\tempurl%
\url{https://doi.org/10.1155/2009/736398}
\showDOI{\tempurl}


\bibitem[\protect\citeauthoryear{Urbanowicz and Moore}{Urbanowicz and Moore}{2015}]%
        {urbanowicz2015exstracs}
\bibfield{author}{\bibinfo{person}{Ryan~John Urbanowicz} {and} \bibinfo{person}{Jason~H Moore}.} \bibinfo{year}{2015}\natexlab{}.
\newblock \showarticletitle{{ExSTraCS} 2.0: description and evaluation of a scalable learning classifier system}.
\newblock \bibinfo{journal}{\emph{Evolutionary intelligence}} \bibinfo{volume}{8}, \bibinfo{number}{2} (\bibinfo{year}{2015}), \bibinfo{pages}{89--116}.
\newblock
\urldef\tempurl%
\url{https://doi.org/10.1007/s12065-015-0128-8}
\showDOI{\tempurl}


\bibitem[\protect\citeauthoryear{Valenzuela-Rend{\'o}n}{Valenzuela-Rend{\'o}n}{1991}]%
        {valenzuela1991fuzzy}
\bibfield{author}{\bibinfo{person}{Manuel Valenzuela-Rend{\'o}n}.} \bibinfo{year}{1991}\natexlab{}.
\newblock \showarticletitle{The fuzzy classifier system: Motivations and first results}. In \bibinfo{booktitle}{\emph{Parallel Problem Solving from Nature}}. \bibinfo{publisher}{Springer Berlin Heidelberg}, \bibinfo{address}{Berlin, Heidelberg}, \bibinfo{pages}{338--342}.
\newblock
\showISBNx{978-3-540-70652-6}
\urldef\tempurl%
\url{https://doi.org/10.1007/BFb0029774}
\showDOI{\tempurl}


\bibitem[\protect\citeauthoryear{Wagner and Stein}{Wagner and Stein}{2022}]%
        {wagner2022mechanisms}
\bibfield{author}{\bibinfo{person}{Alexander R.~M. Wagner} {and} \bibinfo{person}{Anthony Stein}.} \bibinfo{year}{2022}\natexlab{}.
\newblock \showarticletitle{Mechanisms to Alleviate Over-Generalization in {XCS} for Continuous-Valued Input Spaces}.
\newblock \bibinfo{journal}{\emph{SN Computer Science}} \bibinfo{volume}{3}, \bibinfo{number}{2} (\bibinfo{year}{2022}), \bibinfo{pages}{1--23}.
\newblock
\urldef\tempurl%
\url{https://doi.org/10.1007/s42979-022-01060-w}
\showDOI{\tempurl}


\bibitem[\protect\citeauthoryear{Wilson}{Wilson}{1995}]%
        {wilson1995xcs}
\bibfield{author}{\bibinfo{person}{Stewart~W. Wilson}.} \bibinfo{year}{1995}\natexlab{}.
\newblock \showarticletitle{Classifier Fitness Based on Accuracy}.
\newblock \bibinfo{journal}{\emph{Evol. Comput.}} \bibinfo{volume}{3}, \bibinfo{number}{2} (\bibinfo{date}{jun} \bibinfo{year}{1995}), \bibinfo{pages}{149–175}.
\newblock
\showISSN{1063-6560}
\urldef\tempurl%
\url{https://doi.org/10.1162/evco.1995.3.2.149}
\showDOI{\tempurl}


\bibitem[\protect\citeauthoryear{Wilson}{Wilson}{1998}]%
        {wilson1998generalization}
\bibfield{author}{\bibinfo{person}{Stewart~W Wilson}.} \bibinfo{year}{1998}\natexlab{}.
\newblock \showarticletitle{Generalization in the {XCS} Classifier System}.
\newblock \bibinfo{journal}{\emph{Proc. Genetic Programming 1998}} (\bibinfo{year}{1998}).
\newblock


\bibitem[\protect\citeauthoryear{Wilson}{Wilson}{2000}]%
        {wilson1999xcsr}
\bibfield{author}{\bibinfo{person}{Stewart~W. Wilson}.} \bibinfo{year}{2000}\natexlab{}.
\newblock \showarticletitle{Get Real! {XCS} with Continuous-Valued Inputs}. In \bibinfo{booktitle}{\emph{Learning Classifier Systems}}. \bibinfo{publisher}{Springer Berlin Heidelberg}, \bibinfo{address}{Berlin, Heidelberg}, \bibinfo{pages}{209--219}.
\newblock
\showISBNx{978-3-540-45027-6}
\urldef\tempurl%
\url{https://doi.org/10.1007/3-540-45027-0_11}
\showDOI{\tempurl}


\bibitem[\protect\citeauthoryear{Wilson}{Wilson}{2001}]%
        {wilson2000mining}
\bibfield{author}{\bibinfo{person}{Stewart~W. Wilson}.} \bibinfo{year}{2001}\natexlab{}.
\newblock \showarticletitle{Mining Oblique Data with {XCS}}. In \bibinfo{booktitle}{\emph{Advances in Learning Classifier Systems}}. \bibinfo{publisher}{Springer Berlin Heidelberg}, \bibinfo{address}{Berlin, Heidelberg}, \bibinfo{pages}{158--174}.
\newblock
\showISBNx{978-3-540-44640-8}
\urldef\tempurl%
\url{https://doi.org/10.1007/3-540-44640-0_11}
\showDOI{\tempurl}


\bibitem[\protect\citeauthoryear{Wilson}{Wilson}{2002}]%
        {wilson2002classifiers}
\bibfield{author}{\bibinfo{person}{Stewart~W Wilson}.} \bibinfo{year}{2002}\natexlab{}.
\newblock \showarticletitle{Classifiers that approximate functions}.
\newblock \bibinfo{journal}{\emph{Natural Computing}} \bibinfo{volume}{1}, \bibinfo{number}{2} (\bibinfo{date}{04} \bibinfo{year}{2002}), \bibinfo{pages}{211--234}.
\newblock
\urldef\tempurl%
\url{https://doi.org/10.1023/A:1016535925043}
\showDOI{\tempurl}


\bibitem[\protect\citeauthoryear{Wilson}{Wilson}{2008}]%
        {wilson2008classifier}
\bibfield{author}{\bibinfo{person}{Stewart~W. Wilson}.} \bibinfo{year}{2008}\natexlab{}.
\newblock \showarticletitle{Classifier Conditions Using Gene Expression Programming}. In \bibinfo{booktitle}{\emph{Learning Classifier Systems}}. \bibinfo{publisher}{Springer Berlin Heidelberg}, \bibinfo{address}{Berlin, Heidelberg}, \bibinfo{pages}{206--217}.
\newblock
\showISBNx{978-3-540-88138-4}
\urldef\tempurl%
\url{https://doi.org/10.1007/978-3-540-88138-4_12}
\showDOI{\tempurl}


\bibitem[\protect\citeauthoryear{Zadeh}{Zadeh}{1965}]%
        {zadeh1965fuzzy}
\bibfield{author}{\bibinfo{person}{L.A. Zadeh}.} \bibinfo{year}{1965}\natexlab{}.
\newblock \showarticletitle{Fuzzy sets}.
\newblock \bibinfo{journal}{\emph{Information and Control}} \bibinfo{volume}{8}, \bibinfo{number}{3} (\bibinfo{year}{1965}), \bibinfo{pages}{338--353}.
\newblock
\showISSN{0019-9958}
\urldef\tempurl%
\url{https://doi.org/10.1016/S0019-9958(65)90241-X}
\showDOI{\tempurl}


\bibitem[\protect\citeauthoryear{Zadeh}{Zadeh}{1973}]%
        {zadeh1973outline}
\bibfield{author}{\bibinfo{person}{Lotfi~A. Zadeh}.} \bibinfo{year}{1973}\natexlab{}.
\newblock \showarticletitle{Outline of a New Approach to the Analysis of Complex Systems and Decision Processes}.
\newblock \bibinfo{journal}{\emph{IEEE Transactions on Systems, Man, and Cybernetics}} \bibinfo{volume}{SMC-3}, \bibinfo{number}{1} (\bibinfo{date}{Jan} \bibinfo{year}{1973}), \bibinfo{pages}{28--44}.
\newblock
\showISSN{2168-2909}
\urldef\tempurl%
\url{https://doi.org/10.1109/TSMC.1973.5408575}
\showDOI{\tempurl}


\bibitem[\protect\citeauthoryear{Zhang, Stolzenberg-Solomon, Lynch, and Urbanowicz}{Zhang et~al\mbox{.}}{2021}]%
        {zhang2021lcs}
\bibfield{author}{\bibinfo{person}{Robert Zhang}, \bibinfo{person}{Rachael Stolzenberg-Solomon}, \bibinfo{person}{Shannon~M. Lynch}, {and} \bibinfo{person}{Ryan~J. Urbanowicz}.} \bibinfo{year}{2021}\natexlab{}.
\newblock \bibinfo{title}{{LCS-DIVE}: An Automated Rule-based Machine Learning Visualization Pipeline for Characterizing Complex Associations in Classification}.
\newblock
\newblock
\urldef\tempurl%
\url{https://doi.org/10.48550/ARXIV.2104.12844}
\showDOI{\tempurl}


\end{thebibliography}
\end{document}